\documentclass[11pt]{article}

\usepackage[]{acl}

\usepackage{times}
\usepackage{enumitem}
\usepackage{latexsym}
\usepackage{amsmath}
\usepackage{booktabs}
\usepackage{xcolor}
\usepackage{colortbl}
\usepackage[T1]{fontenc}

\usepackage[utf8]{inputenc}

\usepackage{microtype}

\usepackage{inconsolata}

\usepackage{graphicx}
\usepackage[most]{tcolorbox}
\usepackage{tikz}
\usetikzlibrary{shapes.geometric}
\usepackage{pgfplots}
\pgfplotsset{compat=1.18}
\usepgfplotslibrary{groupplots}

\newtcolorbox{examplebox}[1]{
  enhanced,breakable,colback=gray!3,colframe=gray!50,
  boxrule=0.5pt,arc=2pt,left=5pt,right=5pt,top=4pt,bottom=4pt,
  title=\textbf{#1},fonttitle=\small,coltitle=black
}

\usepackage{xcolor}
\usepackage{pifont}

\newcommand{\cmark}{\textcolor{green!60!black}{\ding{51}}}
\newcommand{\xmark}{\textcolor{red!75!black}{\ding{55}}}

\title{\texttt{DuplexSpeechBench-Document Grounding}: Benchmarking Document Grounding and Hallucinations in Voice Agents}

\author{Puneet Mathur\textsuperscript{1}, Nedim Lipka\textsuperscript{1}, Zeyu Jin\textsuperscript{1}, Dinesh Manocha\textsuperscript{2}\\
\textsuperscript{1}\small{Adobe Research, San Jose, USA}\\
\textsuperscript{2}\small{University of Maryland College Park, USA}\\
Project Page: \url{https://dsb-dg.github.io/}}

\begin{document}
\maketitle

\begin{abstract}
Voice agents enable low-latency, natural interaction, yet their ability to faithfully ground responses in external documents remains underexplored. We introduce \textbf{\texttt{DuplexSpeechBench-Document Grounding} (DSB-DG)}, a benchmark for evaluating document grounding in voice agents across five professional domains. DSB-DG targets three failure modes: \textit{Context Saturation}, which measures grounding under increasing document length; \textit{Grounding Decay}, which measures retention of document facts across multi-turn dialogue; and \textit{Proactive Grounding}, which evaluates whether context re-injection mitigates conversational drift. The benchmark contains 1,636 adversarially verified QA pairs from 50 documents covering five professional domains, and supports fully automatic evaluation of grounding accuracy, hallucination, and response latency. Across systems spanning cascaded, proprietary full-duplex and real-time, and open-weight speech2speech architectures, we find substantial differences in effective grounding capacity. While cascaded pipeline (ASR-LLM-TTS) achieves the highest grounding accuracy, Gemini-Live and GPT-Realtime closely trail behind. Open-weight systems exhibit distinct failure modes, most notably an abrupt context-capacity collapse and multi-turn grounding decay. More broadly, grounding fidelity degrades with context and conversational load, and failures frequently manifest as unsupported generations rather than abstention. We show that contextual grounding as a key unresolved challenge for reliable full-duplex voice agents.
\end{abstract}


\begin{figure}[t]
    \centering
    \includegraphics[width=0.49\textwidth]{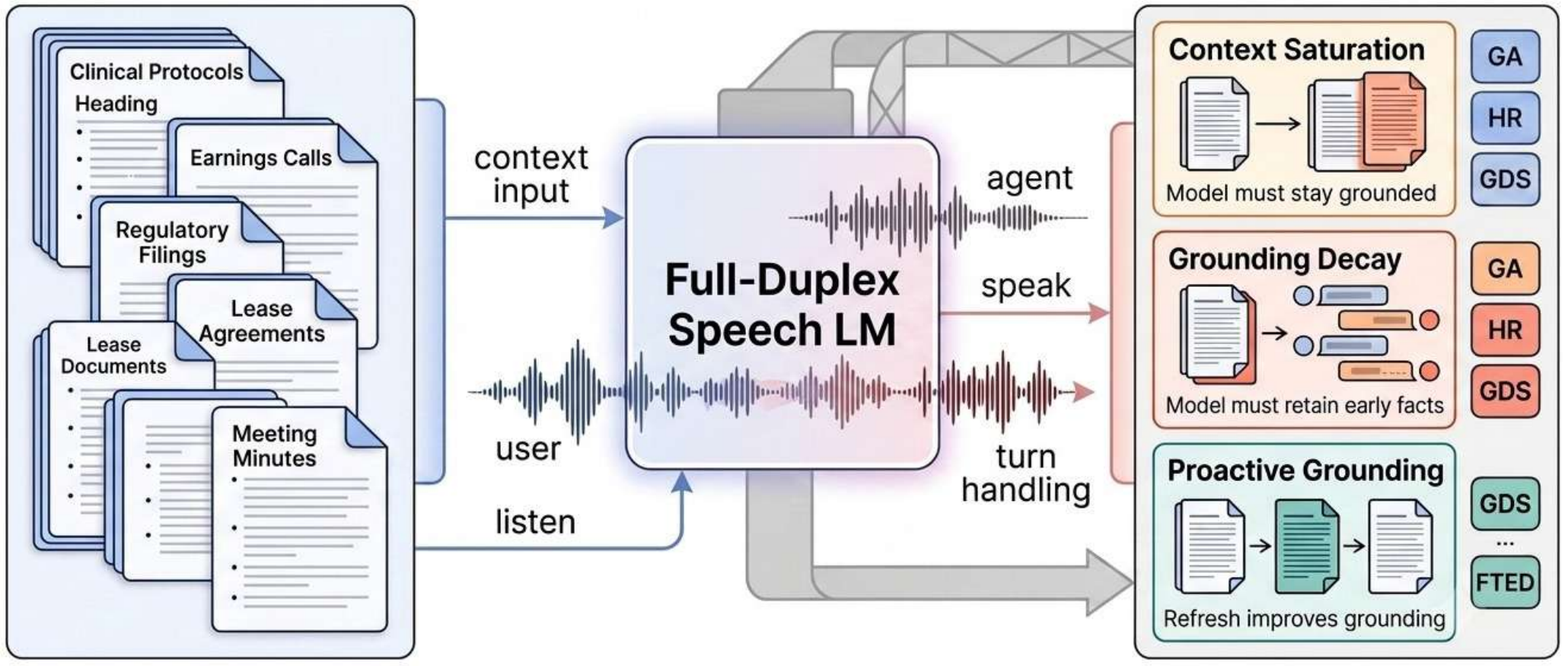}
    \caption{\small{\texttt{DuplexSpeechBench–Document Grounding} (DSB-DG) benchmark provides document context and spoken queries to a full-duplex speech model, and evaluates grounding accuracy, multi-turn retention, and real-time behavior through three complementary protocols.}}
    \label{fig:intro_overview}
\end{figure}

\section{Introduction}
\label{sec:introduction}

Audio language models have rapidly evolved from mere speech transcription systems to full-duplex models capable of simultaneous listening and speaking. Systems such as Moshi~\citep{defossez2024moshi}, GPT-Realtime\footnote{\scriptsize{https://developers.openai.com/api/docs/models/gpt-realtime}}, and MiniCPM-o \cite{cui2026minicpm} enable low-latency interaction with natural turn-taking, interruptions, and backchanneling. As such systems enter professional settings, conversational fluency alone is insufficient: voice agents must also answer faithfully from task-specific documents, such as clinical protocols, financial reports, contracts, or regulatory filings.

\begin{table*}[t]
\centering
\resizebox{0.8\textwidth}{!}{%
\small
\setlength{\tabcolsep}{3.7pt}
\renewcommand{\arraystretch}{1.12}
\begin{tabular}{lcccccccc}
\toprule
\textbf{Benchmark} &
\textbf{Full-} &
\textbf{Real-} &
\textbf{Long} &
\textbf{Long-} &
\textbf{Multi-} &
\textbf{Document-} &
\textbf{Halluc.} &
\textbf{Latency} \\
&
\textbf{Duplex} &
\textbf{Time} &
\textbf{Document} &
\textbf{Context} &
\textbf{Turn} &
\textbf{Factuality} &
\textbf{Eval.} &
\textbf{Eval.} \\
\midrule

SpokenSQuAD~\citep{li2018spoken}
& \xmark & \xmark & \xmark & \xmark & \xmark
& \cmark & \xmark & \xmark \\

LongBench~\citep{bai2024longbench}
& \xmark & \xmark & \cmark & \cmark & \xmark
& \cmark & \xmark & \xmark \\

\midrule

Full-Duplex-Bench~\citep{fdb1,fdbv15}
& \cmark & \cmark & \xmark & \xmark & \xmark
& \xmark & \xmark & \cmark \\

FD-Bench~\citep{fdbench}
& \cmark & \cmark & \xmark & \xmark & \xmark
& \xmark & \xmark & \cmark \\

Full-Duplex-Bench-v2~\citep{fdbv2}
& \cmark & \cmark & \xmark & \xmark & \cmark
& \xmark & \xmark & \cmark \\

MTR-DuplexBench~\citep{mtr}
& \cmark & \cmark & \xmark & \xmark & \cmark
& \xmark & \xmark & \cmark \\

Full-Duplex-Bench-v3~\citep{fdb3}
& \cmark & \cmark & \xmark & \xmark & \cmark
& \xmark & \xmark & \cmark \\

$\tau$-Voice~\citep{tauvoice}
& \cmark & \cmark & \xmark & \xmark & \cmark
& \xmark & \xmark & \cmark \\

\midrule

\textbf{DSB-DG (Ours)}
& \cmark
& \cmark
& \cmark
& \cmark
& \cmark
& \cmark
& \cmark
& \cmark \\

\bottomrule
\end{tabular}
}
\caption{\small{\textbf{Comparison of DSB-DG with spoken-dialogue benchmarks.}
Existing voice agent benchmarks primarily evaluate conversational dynamics,
multi-turn interaction, or task completion, whereas long-context benchmarks
evaluate document understanding without real-time spoken interaction.
DSB-DG jointly evaluates document factuality, multi-turn grounding
retention, hallucination, and latency in full-duplex voice agents.}}
\label{tab:related_work_comparison}
\end{table*}

Document-grounded voice interaction introduces a fundamental challenge where models must preserve factual grounding while operating under the latency and turn-taking constraints of real-time speech. Under contextual load, unsupported responses may be delivered with the same fluency and prosodic confidence as grounded ones. Evaluating when voice agents, including full-duplex and real-time models, retain, lose, or recover access to external context is therefore critical for reliability, yet document grounding in such systems remains largely unexplored.

Existing benchmarks (Table~\ref{tab:related_work_comparison}) address parts of this problem but not their combination. The Full-Duplex-Bench series evaluates turn-taking and interruptions~\citep{fdb1}, overlapping speech~\citep{fdbv15}, multi-turn instruction following~\citep{fdbv2}, and tool use under naturalistic disfluencies~\citep{fdb3}, but not factual fidelity to external documents. Long-context benchmarks such as QuALITY~\citep{pang2022quality}, SCROLLS~\citep{shaham2022scrolls}, and LongBench~\citep{bai2024longbench} evaluate document understanding in text without streaming speech constraints, while SpokenSQuAD~\citep{li2018spoken} focuses on single-turn extractive spoken QA.

\begin{figure*}[t]
    \centering
    \includegraphics[width=0.8\textwidth]{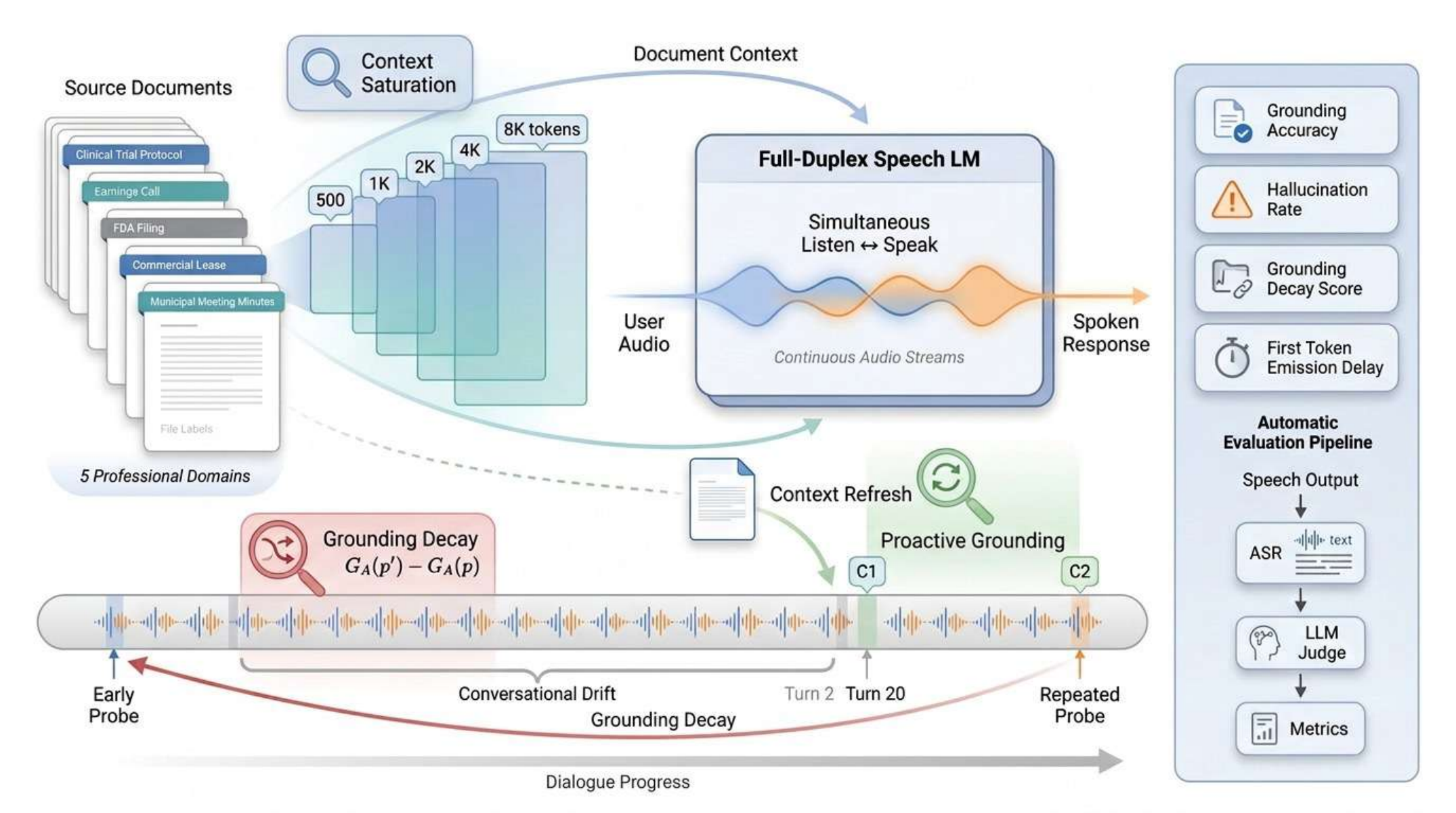}
    \caption{\small{\textbf{DuplexSpeechBench--Document Grounding (DSB-DG)}. We inject document context into a full-duplex speech language model that answers spoken user queries in real time. The benchmark comprises three complementary protocols: \textbf{Context Saturation}, evaluating grounding under increasing document lengths; \textbf{Grounding Decay}, evaluating retention of document facts across multi-turn conversations using repeated factual probes separated by distractor dialogue; and \textbf{Proactive Grounding}, evaluating the effectiveness of periodic document re-injection. Model responses are automatically scored using Grounding Accuracy (GA), Hallucination Rate (HR), Grounding Decay Score (GDS), and First-Token Emission Delay (FTED).}}
    \label{fig:framework_overview}
\end{figure*}

We introduce \textbf{DuplexSpeechBench--Document Grounding (DSB-DG)}, a streaming-native evaluation benchmark for factual grounding and hallucination in document-conditioned voice agents (Figure~\ref{fig:framework_overview}). DSB-DG isolates three complementary failure modes: \textit{Context Saturation}, measures grounding as document length increases; \textit{Grounding Decay}, measures factual retention across extended dialogue; and \textit{Proactive Grounding}, tests whether context re-injection restores grounding after conversational drift. The benchmark contains 1,636 adversarially verified QA pairs over 50 professional documents across five domains---clinical trial protocols, earnings-call transcripts, FDA 510(k) filings, lease agreements, and municipal meeting minutes---with five prefix-consistent context tiers ranging from 500 to 8,000 tokens. All metrics are automatic, enabling reproducible comparison across speech architectures.

We evaluate seven systems spanning full-duplex, cascaded and end-to-end speech architectures: Gemini-Live\footnote{https://gemini.google/overview/gemini-live/}, Cascaded (Whisper $\rightarrow$ GPT-4o $\rightarrow$ TTS), GPT-Realtime, MiniCPM-o-4.5~\citep{cui2026minicpm}, SALMONN~\citep{yu2024salmonn}, Ultravox-v0.6-Qwen3-32B (UV-32B), and Moshi~\citep{defossez2024moshi}. The results reveal distinct architecture-specific failure modes. The cascaded system provides the strongest overall grounding, while Gemini-Live, GPT-Realtime and MiniCPM-o approach its single-turn accuracy under real-time speech generation. SALMONN exhibits an abrupt context-capacity cliff, whereas Ultravox-32B shows severe grounding decay across extended dialogue. Moreover, context re-injection benefits GPT-Realtime and the cascaded system but can degrade smaller-backbone systems, demonstrating that proactive grounding strategies must be architecture-dependent. Across systems, failures under contextual load frequently manifest as unsupported generation rather than abstention, exposing a critical reliability gap between conversational fluency and factual grounding. \textbf{Main contributions}:

\begin{itemize}[nosep]
    \item \textbf{\texttt{DuplexSpeechBench--Document Grounding} Benchmark} for evaluating factual grounding and hallucination in voice agents, comprising 1,636 spoken QA pairs in 50 professional documents spanning 5-tiered context-lengths, along with 20 controlled multi-turn dialogues.

    \item \textbf{Three complementary grounding protocols.}
    \textit{Context Saturation, Grounding Decay, and Proactive Grounding} isolate failures from increasing contextual load, conversational forgetting, and context re-injection using automatic metrics for grounding, hallucination, retention, and latency.

    \item \textbf{Systematic evaluation across speech architectures.}
    Evaluation of cascaded, proprietary full duplex, real-time, and open-weight voice agents reveals distinct failure modes, including context-capacity collapse, conversational forgetting, and context saturation.
\end{itemize}

\noindent We discuss related benchmarks in Appendix \ref{sec:related_work_appendix}. 

\section{Methodology}
\label{sec:methodology}

We evaluate document-grounded voice agents through three controlled protocols that isolate complementary sources of grounding failure. \textbf{Context Saturation} measures how grounding changes as document length increases; \textbf{Grounding Decay} tests whether document facts remain accessible over extended multi-turn interaction; and \textbf{Proactive Grounding} evaluates if periodically reintroducing of document context mitigates conversational drift.

\subsection{Context Saturation}
\label{sec:method_context_saturation}

Context Saturation measures grounding robustness as document length increases. By holding the question and supporting evidence fixed while expanding context, the protocol isolates the effect of contextual load and captures both gradual degradation and abrupt context-capacity collapse.

\noindent \textbf{Setting}: For a source document $D$, we construct five prefix-consistent tiers $T_1 \subset T_2 \subset \cdots \subset T_5$ of approximately 500, 1k, 2k, 4k, and 8k tokens, such that information in $T_i$ remains available in all subsequent tiers. For each question--answer pair $(q,a)$, we ask the same spoken question $q$ under every eligible tier $T_i$ and obtain response $r_i$. Interactions are single-turn with no conversational history, so changes in grounding can be attributed to increasing contextual load.

\begin{table*}[t]
\centering
\resizebox{0.8\textwidth}{!}{%
\small
\begin{tabular}{lcccp{8.5cm}}
\toprule
\textbf{Domain} &
\textbf{\# Docs} &
\textbf{\# QA} &
\textbf{Avg. Qs / Doc} &
\textbf{Document Characteristics} \\
\midrule
Clinical Trials & 10 & 391 & 39.1  &
Structured protocols, eligibility criteria, clinical endpoints \\

Earnings Calls & 10 & 314 & 31.4   &
Financial reporting, analyst Q\&A, dense numerical content \\

FDA 510(k) & 10 & 292 & 29.2  &
Regulatory filings, device specifications, standardized structure \\

Commercial Leases & 10 & 358 & 35.8 &
Contract clauses, cross-references, legal reasoning \\

Municipal Minutes & 10 & 281 & 28.1 &
Procedural narratives, motions, amendments, voting records \\
\midrule
\textbf{Total} & \textbf{50} & \textbf{1,636} & \textbf{32.7} & Multi-turn spoken conversations (20) grounded in documents\\
\bottomrule
\end{tabular}
}
\caption{\small{Domain composition and dataset statistics of \texttt{DuplexSpeechBench-Document Grounding}: 1,636 unique, adversarially verified QA pairs across five distinct professional domains along with complex multi-turn dialogue scripts.}}
\label{tab:domain_stats}
\end{table*}

\subsection{Grounding Decay}
\label{sec:method_grounding_decay}

Context capacity alone does not capture whether document information remains accessible as a conversation progresses. Grounding decay measures retention of document facts under controlled conversational drift.

\noindent \textbf{Setting}: For each document $D$, we construct a multi-turn conversation following a \textit{probe--distractor--repeat} structure. Early turns contain document-grounded probes $A$, intermediate turns introduce domain-relevant distractors $B$, and late turns repeat the original probes as $A'$. As $A'$ is identical to $A$, performance changes reflect conversational progression rather than question variation. We define the Grounding Decay Score: $\mathrm{GDS}=\mathrm{GA}(A')-\mathrm{GA}(A)$, where $\mathrm{GA}(\cdot)$ denotes grounding accuracy over the corresponding probe set. Negative values indicate grounding loss and values near zero indicate stable retention. GDS is interpreted jointly with $\mathrm{GA}(A)$, since near-zero decay is uninformative when initial grounding is poor. Distractors remain domain-related but do not require the probed facts, inducing conversational drift without changing the source context. Conversation templates are provided in Appendix~\ref{app:grounding_decay_protocol}.

\subsection{Proactive Grounding}
\label{sec:method_proactive_grounding}

Proactive Grounding evaluates whether refreshing document context mitigates grounding loss during extended dialogue. Additional context may improve access to source information, but may also overwhelm the effective conversational context of some speech LM architectures.

\noindent \textbf{Setting}: Building on the grounding-decay setup, we compare three context-management conditions with increasing levels of injected context: (i) $\mathbf{C_0}$ (\textbf{No Refresh}): context is provided initially but not re-injected; (ii) $\mathbf{C_1}$ (\textbf{Single Refresh}): context is re-injected once mid-conversation; (iii) $\mathbf{C_2}$ (\textbf{Multiple Refreshes}): context is re-injected multiple times.

\noindent Conversations and factual probes are fixed across conditions; only the refresh strategy changes. We measure re-injection benefit relative to $C_0$ as

{\footnotesize{
\begin{align}
\Delta_{\mathrm{light}} &= \mathrm{GA}(C_1)-\mathrm{GA}(C_0), \\
\Delta_{\mathrm{heavy}} &= \mathrm{GA}(C_2)-\mathrm{GA}(C_0).
\end{align}
}}

\noindent Positive values indicate improved grounding and negative values degradation. See Appendix~\ref{app:proactive_grounding_protocol} for protocol details.


\section{Benchmark Design}
\label{sec:bench}

\paragraph{Benchmark Composition.}
Table~\ref{tab:domain_stats} shows DSB-DG comprises 50 professional documents across five domains: clinical trial protocols, earnings-call transcripts, FDA 510(k) filings, commercial lease agreements, and municipal meeting minutes. The benchmark contains 1,636 verified question--answer pairs and 20 controlled multi-turn conversations. Each document is represented at five prefix-consistent context tiers of approximately 500, 1k, 2k, 4k, and 8k tokens, enabling controlled evaluation of context-length effects while preserving evidence from shorter tiers. Questions span entity and numerical retrieval, procedural reasoning, clause interpretation, and multi-hop document comprehension. Additional dataset statistics in Appendix~\ref{app:dataset_details}.

\paragraph{QA Generation and Verification.} Candidate QA pairs are generated using GPT-OSS-20B paired with restricted document context to ensure the QA is blind to extraneous information, followed by three-stage adversarial verification. This multi-stage process results in 1,636 QA pairs (1,341 grounded and 295 unanswerable). We additionally construct controlled multi-turn conversations containing document-grounded probes, domain-relevant distractors, and topic transitions. See Appendix~\ref{app:qa_verification} for QA verification, Appendix~\ref{app:qa_generation} for prompts, and Appendix~\ref{app:protocol_details} for protocol templates.

\paragraph{Audio Interaction.}
All benchmark queries are synthesized as 16kHz mono speech using a fixed TTS voice, standardizing speaker, prosody, and recording conditions across systems. The benchmark isolates document grounding under standardized acoustic input. TTS processing configurations are provided in Appendix~\ref{app:model_config}.

\paragraph{LLM-Judge Evaluation.}
Model responses are transcribed with Whisper-large-v3~\citep{radford2023robust} and scored by an LLM judge given the source context, question, reference answer, and response transcript to determine if the responses are supported by the supplied document context. See Appendix~\ref{app:judge_validation} for judge evaluation configurations.

\section{Experiments}
\label{sec:experiments}

\noindent\textbf{\underline{Evaluated Systems}}: We evaluate seven spoken-dialogue systems spanning full duplex, cascaded, proprietary real-time, and open-weight speech architectures: \textbf{Gemini-Live}, \textbf{Cascaded} (ASR $\rightarrow$ GPT-4o $\rightarrow$ TTS), \textbf{GPT-Realtime}, \textbf{MiniCPM-o-4.5}~\citep{cui2026minicpm}, \textbf{SALMONN}~\citep{yu2024salmonn}, \textbf{Ultravox-v0.6-Qwen3-32B (UV-32B)}, and \textbf{Moshi}~\citep{defossez2024moshi}. All systems receive identical document content, synthesized user audio, and conversation scripts. See model configurations and API settings in Appendix~\ref{app:model_config}.

\noindent\textbf{\underline{Evaluation Metrics}}: \textbf{Grounding Accuracy (GA)} is the primary quality metric for factual grounding which measures the fraction of responses factually supported by the supplied document. \textbf{Hallucination Rate (HR)} measures fraction of responses containing unsupported claims. For multi-turn evaluation, \textbf{Grounding Decay Score (GDS)} measures the change in accuracy between repeated early and late probes ($\Delta GA$) (Sec. \ref{sec:method_grounding_decay}). Proactive Grounding protocol reports $\Delta_{\mathrm{light}}$ and $\Delta_{\mathrm{heavy}}$ as defined in Section~\ref{sec:method_proactive_grounding}. \textbf{First-Token Emission Delay (FTED)} measures the elapsed time between the end of user speech and the first model output token; we report median (p50), mean, and p95 FTED latencies. We additionally report \textbf{Quality-Adjusted Latency (QAL)} $ = \mathrm{FTED}_{p50}(1-\mathrm{GA_{T3}})$, which penalizes systems that achieve low latency at the expense of grounding accuracy. See Appendix~\ref{sec:context_injection_appendix} for model-specific injection procedures and prompt templates. 



\begin{table}[t]
\centering
\resizebox{0.45\textwidth}{!}{%
\small
\setlength{\tabcolsep}{5.5pt}
\begin{tabular}{lccccc}
\toprule
\textbf{Model} & \textbf{T1 ($\approx$500)} & \textbf{T2 ($\approx$1k)} & \textbf{T3 ($\approx$2k)} & \textbf{T4 ($\approx$4k)} & \textbf{T5 ($\approx$8k)}\\
\midrule
Cascaded              & \textbf{88.7}          & \textbf{89.8}          & \textbf{90.3} & \textbf{90.4} & \textbf{88.5} \\
Gemini-Live & 85.9 & 86.9 & 87.4 & 90.2 & 86.9 \\
GPT-Realtime          & 88.0          & 86.6          & 87.7          & 88.2          & 85.8 \\
MiniCPM-o             & 85.9          & 87.6          & 88.0          & 87.2          & 74.0 \\
UV-32B                & 87.0 & 85.4          & 85.5          & 80.2          & 75.4 \\
SALMONN   & 77.4          & 76.3          & 70.3          & 37.2          & 25.4 \\
Moshi                 & 6.8           & 5.9           & 3.2           & 3.2           & 2.4  \\
\bottomrule
\end{tabular}
}
\caption{\small{Context Saturation, measured via Grounding Accuracy (\%), across token-length tiers. Cascaded, Gemini-Live, and GPT-Realtime stay robust throughout, while MiniCPM-o shows a sharp T5 drop; SALMONN collapses at T4 nears context capacity. Moshi never grounds.}}
\label{tab:exp_a_ga}
\end{table}

\begin{figure}[t]
\centering
\resizebox{0.8\columnwidth}{!}{%
\begin{tikzpicture}
\begin{axis}[
    width=\linewidth,
    height=4.3cm,
    xlabel={Context Tier},
    ylabel={Hallucination Rate (HR \%)},
    xmin=0.8, xmax=5.2,
    ymin=0, ymax=60,
    xtick={1,2,3,4,5},
    xticklabels={T1,T2,T3,T4,T5},
    ytick={0,10,20,30,40,50,60},
    title={\small Hallucination Rate (\%) Under Context Load},
    grid=major,
    grid style={dotted, gray!50},
    tick style={draw=none},
    yticklabel style={font=\small},
    xticklabel style={font=\small},
    xlabel style={font=\small},
    ylabel style={font=\small},
    every axis plot/.append style={line width=0.8pt},
    clip=false,
    legend style={
        font=\scriptsize,
        at={(0.5,-0.30)},
        anchor=north,
        draw=none,
        fill=none,
        legend columns=3,
        column sep=4pt,
        row sep=1pt,
    }
]

\addplot[
    color=black,
    mark=square*,
    mark size=2pt
] coordinates {
    (1,12.2)
    (2,10.6)
    (3,9.5)
    (4,9.5)
    (5,12.3)
};
\addlegendentry{Cascaded}

\addplot[
    color=blue,
    mark=triangle*,
    mark size=2pt
] coordinates {
    (1,11.4)
    (2,14.2)
    (3,11.8)
    (4,12.1)
    (5,14.8)
};
\addlegendentry{GPT-RT}

\addplot[
    color=magenta!80!black,
    mark=star,
    mark size=3.2pt,
    line width=1.0pt,
    mark options={
        fill=magenta!70,
        draw=black,
        line width=0.6pt
    }
] coordinates {
    (1,13.3)
    (2,13.4)
    (3,12.7)
    (4,10.5)
    (5,13.7)
};
\addlegendentry{Gemini-Live}

\addplot[
    color=teal,
    mark=*,
    mark size=2pt
] coordinates {
    (1,13.9)
    (2,12.7)
    (3,12.3)
    (4,13.1)
    (5,25.0)
};
\addlegendentry{MiniCPM-o}

\addplot[
    color=orange,
    mark=diamond*,
    mark size=2pt
] coordinates {
    (1,13.0)
    (2,11.6)
    (3,11.4)
    (4,10.5)
    (5,22.7)
};
\addlegendentry{UV-32B}

\addplot[
    color=red,
    mark=x,
    mark size=3pt,
    dashed
] coordinates {
    (1,22.6)
    (2,24.4)
    (3,30.1)
    (4,39.8)
    (5,46.5)
};
\addlegendentry{SALMONN}

\draw[red, dashed, thin]
    (axis cs:3.5,0) --
    (axis cs:3.5,60)
    node[below left, font=\tiny, red] {cliff};

\end{axis}
\end{tikzpicture}%
}
\caption{\small{Hallucination Rate (HR) across context tiers. SALMONN's T3$\to$T4 collapse coincides with rising hallucination, while MiniCPM-o shows a similar failure at T5. Context-capacity limits manifest as unsupported generation.}}
\label{fig:joint_grounding_analysis}
\end{figure}
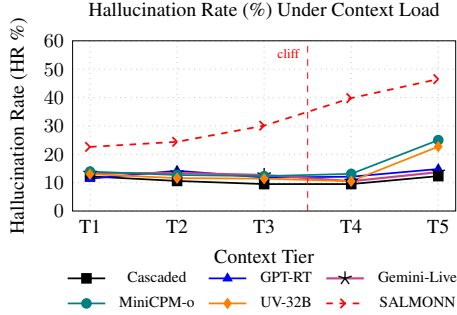

\begin{table}[t]
\centering
\scalebox{0.56}{%
\setlength{\tabcolsep}{3.8pt}
\begin{tabular}{lcccccc}
\toprule
\textbf{Domain} & \textbf{Cascaded} & \textbf{GPT-RT} & \textbf{Gemini-L} & \textbf{MiniCPM-o} & \textbf{UV-32B} & \textbf{SALMONN} \\
\midrule

Clinical
& \cellcolor{red!24}\textbf{83.9}
& \cellcolor{red!30}78.9
& \cellcolor{red!24}81.4
& \cellcolor{red!24}80.7
& \cellcolor{red!30}82.5
& \cellcolor{red!24}62.5 \\

Earnings
& \cellcolor{red!18}\textbf{93.8}
& \cellcolor{red!18}89.4
& \cellcolor{red!18}88.2
& \cellcolor{red!18}91.9
& \cellcolor{red!18}91.9
& \cellcolor{red!18}70.8 \\

FDA 510(k)
& \cellcolor{red!30}82.1
& \cellcolor{red!24}\textbf{84.6}
& \cellcolor{red!30}78.6
& \cellcolor{red!30}78.6
& \cellcolor{red!24}\textbf{84.6}
& \cellcolor{red!30}57.3 \\

Leases
& \cellcolor{red!6}\textbf{98.6}
& \cellcolor{red!6}\textbf{98.6}
& \cellcolor{red!6}96.4
& \cellcolor{red!12}95.7
& \cellcolor{red!6}97.8
& \cellcolor{red!6}86.2 \\

Municipal
& \cellcolor{red!12}\textbf{96.1}
& \cellcolor{red!12}93.3
& \cellcolor{red!12}95.0
& \cellcolor{red!6}\textbf{96.1}
& \cellcolor{red!12}95.0
& \cellcolor{red!12}78.2 \\

\bottomrule
\end{tabular}
}
\caption{\small{Grounding Accuracy (\%) at the 2k-token tier (T3) across professional domains. Darker red cells reflect relative domain difficulty within each model family; bold is best model for each domain. Clinical protocols and FDA 510(k) filings are most challenging domains, while leases and municipal records are most tractable.}}
\label{tab:exp_a_domain}
\end{table}
\begin{table}[t]
\centering
\small
\scalebox{0.8}{%
\setlength{\tabcolsep}{5pt}
\begin{tabular}{lcccc}
\toprule
\textbf{Model} & \textbf{GA$_A$} & \textbf{GA$_B$} & \textbf{GA$_{A'}$} & \textbf{GDS} \\
\midrule
Gemini-Live   & 86.7 & 85.1 & 90.1 & \cellcolor{green!6}{$\mathbf{+3.3}$} \\
Cascaded      & 86.7 & 86.7 & 86.7 & $+$0.0 \\
GPT-Realtime  & 86.7 & 85.0 & 87.0 & $+$0.3 \\
MiniCPM-o     & 83.3 & 86.7 & 81.7 & $-$1.7 \\
UV-32B        & 80.0 & 78.3 & 75.0 & \cellcolor{red!6}{$\mathbf{-5.0}$} \\
SALMONN       & 76.7 & 65.0 & 72.1 & $-$4.6 \\
Moshi         & 1.7  & 5.0  & 1.3  & $-$0.4 \\
\bottomrule
\end{tabular}
}
\caption{\small{Grounding Decay results across controlled multi-turn conversations. GA (\%) is reported for early probes ($A$), intermediate probes ($B$), and repeated late probes ($A'$), with GDS$\,=\,$GA$(A')-$GA$(A)$. Gemini-Live is most robust. UV-32B exhibits most conversational forgetting ($-$ve GDS).}
\label{tab:exp_b}}
\end{table}

\section{Results}
\label{sec:results} 
\paragraph{Context Saturation Analysis.} Table~\ref{tab:exp_a_ga} and Figure~\ref{fig:joint_grounding_analysis} show substantially different context-scaling behavior across architectures. Cascaded, GPT-Realtime, Gemini-Live, and MiniCPM-o remain comparatively stable through T4, achieving 90.3\%, 87.7\%, 87.4\%, and 88.0\% GA at T3, respectively. Gemini-Live remains robust across context tiers, ranging from 85.9\% at T1 to 90.2\% at T4 and retaining 86.9\% at T5. UV-32B degrades more gradually from 85.5\% at T3 to 75.4\% at T5, whereas SALMONN exhibits an abrupt context-capacity cliff: GA falls from 70.3\% at T3 to 37.2\% at T4 ($-33.1$ \%). MiniCPM-o shows a later drop at T5, falling from 88.0\% at T3 to 74.0\%. Moshi remains near floor-level grounding across all standard context tiers ($\leq$6.8\%). Thus, increasing document length produces qualitatively different failure regimes across systems rather than a common degradation curve.

\paragraph{Capacity limits amplify hallucination.} Grounding degradation is accompanied by increasing unsupported generation (Figure~\ref{fig:joint_grounding_analysis}). SALMONN's T3$\rightarrow$T4 accuracy collapse coincides with HR increasing from 30.1\% to 39.8\%, reaching 46.5\% at T5. MiniCPM-o and UV-32B show the same pattern at T5, with HR increasing to 25.0\% and 22.7\%, respectively. In contrast, Cascaded, GPT-Realtime, and Gemini-Live remain comparatively stable; Gemini-Live maintains HR between 10.5\% and 13.7\% across all five tiers. Thus, context-capacity failures often manifest as hallucination rather than abstention.

\paragraph{Document domain also affects grounding.} Table~\ref{tab:exp_a_domain} shows that clinical trial protocols and FDA 510(k) filings are consistently the most challenging due to their dense, specification-heavy content and need for fine-grained factual distinctions. In contrast, commercial leases and municipal minutes remain the most robust domains, with GA ranging from 93.3\% to 98.6\% across Cascaded, GPT-Realtime, Gemini-Live, MiniCPM-o, and UV-32B. SALMONN is weakest on FDA filings (57.3\%) and clinical protocols (62.5\%), but strongest on leases (86.2\%). Grounding robustness depends on both context length and document domain.

\begin{table}[t]
\centering
\resizebox{0.5\textwidth}{!}{%
\small
\setlength{\tabcolsep}{6pt}
\begin{tabular}{lccc|cc}
\toprule
& \multicolumn{3}{c}{\textbf{Grounding Accuracy (\%)}} 
& \multicolumn{2}{c}{\textbf{Injection Benefit (\%)}} \\
\cmidrule(lr){2-4}
\cmidrule(lr){5-6}
\textbf{Model}
& $\mathbf{C_0}$
& $\mathbf{C_1}$
& $\mathbf{C_2}$
& $\boldsymbol{\Delta}_{\text{light}}$
& $\boldsymbol{\Delta}_{\text{heavy}}$ \\
&
\textit{No refresh}
& \textit{1 refresh}
& \textit{Multiple}
& $(C_1-C_0)$
& $(C_2-C_0)$ \\
\midrule
GPT-Realtime & 79.6 & 94.2 & \textbf{88.1} & $+$14.6 & $+$8.5 \\
Cascaded & 87.8 & 94.2 & \textbf{94.9} & $+$6.4 & $+$7.1 \\
Gemini-Live & 83.89 & \textbf{84.44} & 83.89 & $+$0.56 & 0.0 \\
MiniCPM-o     & \textbf{90.6} & 89.4 & 89.4 & $-$1.1 & $-$1.1 \\
UV-32B        & \textbf{88.9} & 88.3 & 86.7 & $-$0.6 & $-$2.2 \\
SALMONN       & \textbf{68.9} & 66.1 & 64.4 & $-$2.8 & $-$4.4 \\
Moshi         & \textbf{5.0}  & 3.3  & 1.7  & $-$1.7 & $-$3.3 \\
\bottomrule
\end{tabular}
}
\caption{\small{Grounding Accuracy (GA) under Proactive Grounding protocol: no refresh ($C_0$), a single refresh ($C_1$), and multiple refreshes ($C_2$), with improvement relative to $C_0$. Cascaded and GPT-Realtime benefit substantially from context re-injection, whereas repeated refreshes degrade UV-32B, MiniCPM-o, and
SALMONN. Bold indicates the best GA.}}
\label{tab:exp_c}
\end{table}



\begin{table}[t]
\centering
\resizebox{0.5\textwidth}{!}{%
\small
\setlength{\tabcolsep}{5.5pt}
\begin{tabular}{lrrrr}
\toprule
\textbf{Model} & \textbf{p50 (ms) $\downarrow$} & \textbf{mean (ms) $\downarrow$} & \textbf{p95 (ms) $\downarrow$} & \textbf{QAL (ms) $\downarrow$} \\
\midrule
Cascaded       & 1469 & 8118 & 61563 &  \textbf{143} \\
GPT-Realtime   & 1214 & 1311 &  1761 &  150 \\
Moshi          &     229 &    251 &      737 & 228 \\
MiniCPM-o      & 2317 & 4566 & 11975 &  278 \\
Gemini-Live    & 6806 & 9410 & 6949 & 856 \\
SALMONN        & 3763 & 7665 & 30349 & 1118  \\
UV-32B         & 9855 & 10723 & 21530 & 1036 \\
\bottomrule
\end{tabular}
}
\caption{\small{First-Token Emission Delay (FTED) statistics and Quality-Adjusted Latency (QAL\,$=$\,FTED$_{\text{p50}}\times(1-\mathrm{GA@T3})$) across systems. GPT-Realtime and the cascaded pipeline dominate the frontier. Moshi attains has lowest latency but near-zero grounding. MiniCPM-o, UV-32B, and SALMONN exhibit severe latency tails at higher context tiers.}}
\label{tab:fted}
\end{table}

\begin{figure}[t]
\centering
\resizebox{0.8\linewidth}{!}{%
\begin{tikzpicture}
\definecolor{cCascaded}{HTML}{2166AC}
\definecolor{cGPT}{HTML}{4DAC26}
\definecolor{cMiniCPM}{HTML}{D7191C}
\definecolor{cSALMONN}{HTML}{F4A582}
\definecolor{cUV}{HTML}{B2ABD2}
\definecolor{cMoshi}{HTML}{969696}
\definecolor{cGemini}{HTML}{FF7F00}
\definecolor{cFront}{HTML}{888888}

\begin{axis}[
    width=1.0\columnwidth,
    height=6.2cm,
    xmode=log,
    log basis x=10,
    xmin=180,
    xmax=13000,
    ymin=-0.04,
    ymax=1.05,
    xtick={200,500,1000,2000,5000,10000},
    xticklabels={200,500,{1000},{2000},{5000},{10000}},
    ytick={0,0.2,0.4,0.6,0.8,1.0},
    yticklabel={
        \pgfmathparse{100*\tick}
        \pgfmathprintnumber[precision=0]{\pgfmathresult}\%
    },
    scaled ticks=false,
    xlabel={FTED p50 (ms, log scale)},
    ylabel={Grounding Accuracy --- GA@T3},
    tick label style={font=\scriptsize},
    label style={font=\scriptsize},
    grid=both,
    major grid style={dashed, gray!30},
    minor grid style={dotted, gray!15},
    clip=true
]

\addplot[
    const plot,
    color=cFront,
    line width=1.2pt,
    dashed
] coordinates {
    (229, 0.032) (1214, 0.877) (1469, 0.903) (13000, 0.903)
};

\addplot[
    only marks,
    mark=diamond*,
    mark size=3.8pt,
    mark options={fill=cCascaded, draw=black, line width=0.9pt}
] coordinates {(1469,0.903)};

\addplot[
    only marks,
    mark=square*,
    mark size=2.8pt,
    mark options={fill=cGPT, draw=black!55, line width=0.6pt}
] coordinates {(1214,0.877)};

\addplot[
    only marks,
    mark=*,
    mark size=2.8pt,
    mark options={fill=cMiniCPM, draw=black!55, line width=0.6pt}
] coordinates {(2317,0.880)};

\addplot[
    only marks,
    mark=pentagon*,
    mark size=3.0pt,
    mark options={fill=cUV, draw=black!55, line width=0.6pt}
] coordinates {(9855,0.855)};

\addplot[
    only marks,
    mark=triangle*,
    mark size=3.8pt,
    mark options={fill=cSALMONN, draw=black, line width=0.9pt}
] coordinates {(3763,0.703)};

\addplot[
    only marks,
    mark=x,
    mark size=4.2pt,
    mark options={draw=cMoshi, line width=1.5pt}
] coordinates {(229,0.032)};

\node[
    regular polygon,
    regular polygon sides=6,
    minimum size=4.2pt,
    inner sep=0pt,
    fill=violet,
    draw=black,
    line width=0.6pt
] at (axis cs:6806,0.884) {};

\node[font=\tiny\itshape, color=black!45, anchor=north west]
    at (rel axis cs:0.04,0.96) {upper-left = better};

\node[font=\tiny, anchor=west]
    at (axis cs:240,0.085) {Moshi};

\node[font=\tiny, anchor=east]
    at (axis cs:1120,0.80) {GPT-Realtime};

\node[font=\tiny, anchor=south]
    at (axis cs:1469,0.94) {Cascaded};

\node[font=\tiny, anchor=west]
    at (axis cs:2600,0.92) {MiniCPM-o};

\node[font=\tiny, anchor=east]
    at (axis cs:3450,0.76) {SALMONN};

\node[font=\tiny, anchor=east]
    at (axis cs:9400,0.85) {UV-32B};

\node[font=\tiny, anchor=east]
    at (axis cs:6500,0.88) {Gemini-Live};

\end{axis}
\end{tikzpicture}
}
\caption{\small{Pareto frontier graph of the quality--latency trade-off in DSB-DG.}}
\label{fig:latency_tradeoff_single}
\label{pareto}
\end{figure}
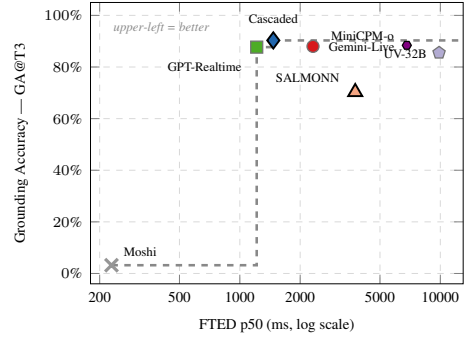


\paragraph{Grounding Decay Analysis.} Table~\ref{tab:exp_b} shows that strong single-turn grounding does not always persist across dialogue. Gemini-Live exhibits the strongest retention with GDS of $=+3.3\%$. Cascaded remains perfectly stable at 86.7\% (GDS $=0.0$ \%), while GPT-Realtime changes only slightly from 86.7\% to 87.0\% ($+0.3$ \%). In contrast, UV-32B shows the largest decay, from 80.0\% to 75.0\% (GDS $=-5.0$ \%), followed by SALMONN (76.7\%$\rightarrow$72.1\%, $-4.6$ \%) and MiniCPM-o (83.3\%$\rightarrow$81.7\%, $-1.7$ \%). Moshi's near-zero GDS ($-0.4$ \%) is uninformative given its near-zero initial grounding (GA$_A=1.7\%$). Thus, GDS must be interpreted jointly with absolute GA: low decay indicates retention only when grounding is established initially. See Appendix~\ref{app:additional_results} for conversation trajectories and failure examples.


\paragraph{Proactive Grounding Analysis.} Table~\ref{tab:exp_c} shows that context re-injection has strongly architecture-dependent effects. Cascaded improves from 87.8\% to 94.9\% GA with multiple refreshes ($\Delta_{\mathrm{heavy}}=+7.1$ \%), while GPT-Realtime rises from 79.6\% to 94.2\% with one refresh ($\Delta_{\mathrm{light}}=+14.6$ \%) and remains improved with multiple refreshes (88.1\%, $+8.5$ \%). Gemini-Live is largely insensitive to re-injection. In contrast, repeated refreshes reduce GA for UV-32B ($-2.2$ \%), MiniCPM-o ($-1.1$ \%), and SALMONN ($-4.4$ \%). Thus, proactive grounding is not uniformly beneficial: its effect depends on both architecture and refresh frequency.




\paragraph{Quality--Latency Trade-off.} Table~\ref{tab:fted} and Figure~\ref{pareto} characterize the quality--latency trade-off across systems. Cascaded achieves the strongest grounding (90.3\%) with a p50 FTED of 1469\,ms and QAL of 143, while GPT-Realtime combines 87.7\% GA with a p50 FTED of 1214\,ms and QAL of 150. MiniCPM-o is the strongest open-weight system in grounding (88.0\%) but has higher latency (2317\,ms p50). Gemini-Live achieves similarly strong grounding (87.4\%) but at substantially higher latency (6806\,ms p50 and QAL of 856). SALMONN exhibits a heavy latency tail (3763\,ms p50; 30349\,ms p95), while UV-32B reaches 9855\,ms p50. Moshi is fastest (229\,ms p50) but provides near-zero grounding (3.2\% GA). Thus, low latency alone is insufficient, while high grounding accuracy alone does not guarantee an effective real-time system: voice agents must be evaluated jointly on responsiveness and factual grounding. 


\section{Conclusion}
\label{sec:conclusion}

We introduced \texttt{DuplexSpeechBench-Document Grounding} (DSB-DG) benchmark for evaluating document grounding in voice agents. We identify distinct failure modes under contextual and conversational load: abrupt context-capacity collapse, multi-turn grounding decay, and increased hallucination as grounding deteriorates. Our findings show that conversational responsiveness alone is insufficient for reliable document-grounded voice interactions. DSB-DG provides a reproducible benchmark for developing full-duplex voice agents that jointly optimize factual grounding and conversational latency.

\section{Limitations}
\label{sec:limitations}

DSB-DG is designed to isolate document grounding under controlled spoken interaction, and several limitations define the scope of its conclusions. First, the benchmark covers five professional domains and contexts up to approximately 8k tokens; substantially longer documents, other domains, and less structured knowledge sources remain outside the present evaluation. Second, user questions are rendered using standardized synthetic speech. This controls acoustic variability across systems but does not measure robustness to speaker diversity, accents, background noise, spontaneous speech, or naturally occurring disfluencies. Third, the evaluated architectures expose different mechanisms for conditioning on external text. We use each system's closest available context interface, but these mechanisms are not architecturally equivalent. The results therefore characterize complete deployed systems rather than a controlled comparison of language-model backbones alone.  Fourth, grounding quality is evaluated primarily with an LLM-based judge. The structured judge captures factual correctness and hallucination behavior at scale, but residual errors may remain for partially correct, implicitly supported, or acoustically ambiguous responses. Finally, DSB-DG focuses on factual grounding, conversational retention, context re-injection, and response latency rather than the complete capability profile of production voice agents. It does not directly evaluate naturalness, emotion, speaker similarity, interruption handling, safety, or task-completion quality. These dimensions are complementary to document grounding and should be considered jointly in real-world deployment.

\section{Ethics Statement}

DSB-DG is designed to evaluate document grounding in full-duplex speech systems without relying on personal or sensitive user information. The benchmark does not contain private conversations, personally identifying information, or person-specific factual queries, and is not intended to retrieve or infer sensitive attributes about individuals. Documents are used solely as grounding sources for controlled evaluation, and benchmark questions target document content rather than personal profiling. All user speech is synthetically generated using a fixed TTS voice, avoiding the collection or release of human voice recordings and reducing privacy risks associated with speaker identity. Model outputs may nevertheless contain hallucinated or unsupported information, particularly under long-context or multi-turn conditions; the benchmark is intended to measure these failure modes rather than endorse deployment of the evaluated systems in high-stakes settings. Results should therefore not be interpreted as evidence that any evaluated system is suitable for autonomous use in domains such as healthcare, finance, legal services, or regulatory decision-making without additional safeguards and human oversight.

\bibliography{custom}
\clearpage
\newpage

\appendix
\section*{Appendix}
\label{sec:appendix}

This appendix provides additional details on benchmark construction,
automatic evaluation, experimental configurations, evaluation protocols,
and supplementary analyses for
\texttt{DuplexSpeechBench-Document Grounding} (DSB-DG).


\section{Related Work}
\label{sec:related_work_appendix}

\paragraph{Real-Time Voice Agent Benchmarks.}
Recent benchmarks for full-duplex voice agents primarily evaluate conversational dynamics. The \texttt{Full-Duplex-Bench} series measures turn-taking, pause and interruption handling~\citep{fdb1}, overlapping speech and backchanneling~\citep{fdbv15}, multi-turn instruction following with an automated examiner~\citep{fdbv2}, and multi-step tool use under naturalistic disfluencies~\citep{fdb3}. MTR-DuplexBench further evaluate dialogue quality, instruction following, safety, and consistency across extended interactions~\citep{mtr}. While these benchmarks provide critical frameworks for auditing the temporal fluidness and conversational mechanics of full-duplex models,  they do not measure a model's fidelity to external, long-context knowledge sources or characterize the acoustic-modality degradation that occurs under heavy contextual load. Our proposed DSB-DG complements this line of work by evaluating whether full-duplex agents can retrieve and preserve document-grounded information while maintaining real-time spoken interaction.

\paragraph{Spoken Question Answering.}
Prior work at the intersection of speech processing and information retrieval has primarily focused on extractive spoken question answering (SQA). Datasets such as SpokenSQuAD \citep{li2018spoken} and its derivatives evaluate a system's capacity to locate exact answer spans within audio documents or automated transcriptions. However, these resources are fundamentally limited to single-turn, short-context scenarios, typically bounding the target audio to brief paragraphs or isolated audio segments. They do not account for multi-turn dialogue drift, long-range structural dependencies, or the open-weight full-duplex architectures that process continuous audio tokens natively.

\paragraph{Long-Context Document Grounding and RAG.}
Long-context NLP benchmarks including QuALITY~\citep{pang2022quality}, SCROLLS~\citep{shaham2022scrolls}, and LongBench~\citep{bai2024longbench} evaluate reasoning and information retrieval over extended textual contexts. Subsequent work has characterized specific long-context failure modes, including ``lost in the middle'' effects~\citep{liu2024lost} and degradation under hard negatives~\citep{jin2025long}, while benchmarks such as Long$^2$RAG~\citep{long2rag} and M-LongDoc~\citep{chia-etal-2025-longdoc} extend evaluation to long-form retrieval and multimodal documents. These benchmarks provide strong foundations for studying long-context grounding, but assume static text-based interaction and therefore do not capture the latency, conversational drift, or grounding persistence of real-time speech agents.

\noindent DSB-DG lies at the intersection of these research directions. Unlike full-duplex benchmarks, it evaluates factual fidelity to external documents; unlike SQA, it targets long-context, generative, multi-turn interaction; and unlike text-based long-context benchmarks, it evaluates grounding within a real-time voice agents loop. Its three protocols further separate failures caused by increasing document length, conversational forgetting, and context re-injection, enabling controlled analysis of document grounding in full-duplex speech systems.

\section{Dataset and Benchmark Construction}
\label{app:dataset_details}

\subsection{Document Sources and Selection}
\label{app:document_sources}

DSB-DG contains 50 professional documents, with 10 documents from each of
five domains: clinical trial protocols, earnings-call transcripts, FDA
510(k) filings, commercial lease agreements, and municipal meeting minutes.
Documents were obtained from the following public sources
clinical trial protocols from the ClinicalTrials.gov API v2 (\texttt{clinicaltrials.gov/api/v2/studies}); earnings-call transcripts from SEC EDGAR full-text search (\texttt{efts.sec.gov}); FDA 510(k) clearance summaries from the openFDA device API (\texttt{api.fda.gov/device/510k.json}); municipal meeting minutes from the Legistar WebAPI for seven U.S. cities (Chicago, Seattle, Boston, Denver, Phoenix, Minneapolis, Portland); and commercial lease agreements that are \emph{synthetically generated} (no external source). Documents were selected according to per-domain retrieval filters with a target of 10 documents each: sector/SIC quotas (e.g.\ five clinical sub-areas at 2 docs each; five FDA device sectors at 2 each), date windows (clinical from 2024-01-01; earnings and FDA 2024-01-01 to 2026-03-31; municipal from 2024-06-01), completed-study/decision-code status
filters (clinical \texttt{status=COMPLETED}; FDA \texttt{decision\_code=SESE}), and keyword/length filters (municipal requires deliberative keywords such as ``motion/vote/ayes/nays/resolution/approved'' and $\geq$2000 characters) and excluded when a document failed these filters. For earnings, a fixed set of mega-cap tickers was explicitly excluded (AAPL, GOOGL, MSFT, AMZN, META, NVDA, TSLA). Table~\ref{tab:document_sources} summarizes the provenance and basic
statistics of the corpus.

\begin{table*}[t]
\centering
\small
\setlength{\tabcolsep}{4pt}
\begin{tabular}{lclrr}
\toprule
\textbf{Domain} &
\textbf{\# Docs} &
\textbf{Source} &
\textbf{Mean Tokens} &
\textbf{Token Range} \\
\midrule
Clinical Trials
& 10 & ClinicalTrials.gov & 2,528 & 1,244--8,038 \\

Earnings Calls
& 10 & SEC EDGAR & 9,403 & 1,370--22,524 \\

FDA 510(k)
& 10 & openFDA & 4,946 & 2,788--7,301 \\

Commercial Leases
& 10 & Synthetic & 6,779 & 6,671--6,875 \\

Municipal Minutes
& 10 & Legistar & 3,092 & 1,335--7,660 \\
\bottomrule
\end{tabular}
\caption{Document provenance and length statistics for the five DSB-DG
domains. Mean/range are full-document token counts
(\texttt{source\_tokens}, before tiering). Token counts are computed using
the \texttt{cl100k\_base} encoding (\texttt{tiktoken}).}
\label{tab:document_sources}
\end{table*}

\subsection{Document Preprocessing}

Raw documents are converted to text using domain-specific parsers that retrieve structured text directly from each provider API (ClinicalTrials.gov, SEC EDGAR, openFDA, Legistar) or, for commercial leases, generate the text programmatically; no PDF rendering is involved. We remove provider API/markup boilerplate during parsing
while preserving the narrative document body used for QA generation. No OCR was required (all sources provide machine-readable text). No document content is summarized or reordered before context-tier construction.

\subsection{Context-Tier Construction}
\label{app:context_saturation_protocol}

Documents are tokenized using the \texttt{cl100k\_base} encoding (\texttt{tiktoken} 0.7.0). We construct five target context tiers of approximately 500, 1k, 2k, 4k, and 8k tokens using a content-preserving budget allocation. The tiers are prefix-consistent such that information present at tier $T_i$ remains available in all subsequent tiers. No text is reordered or summarized between tiers.

\subsection{QA Generation}
\label{app:qa_generation}

Candidate question--answer pairs are generated using GPT-OSS-20B. The generation model receives only the plain text of the single assigned tiered document (no metadata, no other tiers) to prevent access to information outside the assigned source context. Generation uses temperature 0.7, maximum output length 4096 tokens, and a fixed JSON-only output contract. For each document-tier pair, we generate 12 candidates. Candidate questions target the following reasoning categories: \textit{entity retrieval}, \textit{numerical retrieval}, \textit{procedural reasoning}, \textit{clause interpretation}, and \textit{multi-hop document comprehension}. The question taxonomy is encoded as grounded (subtypes \texttt{direct\_lookup}, \texttt{multi\_span}, \texttt{inferential}) and unanswerable (subtypes \texttt{adjacent}, \texttt{temporal}, \texttt{overshoot}). The reasoning categories are classified as entity, numerical, clause, procedural, multi-hop).

\paragraph{QA Generation Prompt.} The generation prompt used in our experiments is reproduced below.

\begin{quote}
\small
\ttfamily
[SYSTEM]\\
You are a benchmark data construction expert.\\
Given a document, generate QA pairs for a grounding evaluation benchmark.\\
RULES:\\
1. Every answer MUST be explicitly stated --- no inference, no math.\\
2. Copy EXACT sentences from the document as supporting\_sentences.\\
3. Questions sound like natural spoken questions --- no ``According to
paragraph 3'' style.\\
4. Unanswerable questions must be genuinely unanswerable from this document.\\
5. Output ONLY valid JSON, no preamble, no markdown code fences.\\
6. Wrap the array in a JSON object: \{``qa\_pairs'': [...]\}\\[4pt]
[USER]\\
Document:\\
"""\{document\}"""\\[2pt]
Generate:\\
- 8 grounded (subtype: direct\_lookup | multi\_span | inferential)\\
- 4 unanswerable (subtype: adjacent | temporal | overshoot)\\[2pt]
JSON schema per item in the qa\_pairs array:\\
\{ question\_text, question\_type (grounded|unanswerable),
question\_subtype, ground\_truth\_answer, supporting\_sentences [exact
verbatim sentence from doc], is\_answerable, difficulty (easy|medium|hard),
document\_position (early|middle|late) \}\\[2pt]
Return JSON object: \{``qa\_pairs'': [list of QA objects]\}.
\end{quote}

\subsection{QA Verification Pipeline}
\label{app:qa_verification}

All generated QA pairs undergo three verification stages.

\paragraph{Stage 1: Answerability.} Each candidate is evaluated to ensure that its reference answer is explicitly supported by the assigned document context. Each instance is checked by an automated answerability filter following the Stage-2 deterministic criteria: the answer must be $\leq$15 words for \texttt{direct\_lookup} items and $\geq$2 characters, the question must end with ``?'', and at least one \texttt{supporting\_sentence} must match the document under whitespace-normalized exact match. A candidate is retained when it passes all deterministic checks (automated; no voting).

\paragraph{Stage 2: Ambiguity.} All questions are checked for alternative plausible answers, underspecified references, and conflicting evidence. Verification is
performed using GPT-OSS-120B. Unanswerable QA candidates are rejected if their keyword overlap with the evidence text exceeds 0.60 (or if it could be answered from the question alone). Questions with more than one defensible answer are removed.

\paragraph{Stage 3: Cross-Tier Consistency.} For every QA pair assigned to $T_i$, we verify that all evidence required for its reference answer occurs within $T_i$ based on the location of the reference answer from the document. Prefix consistency then ensures that the question remains answerable at every subsequent tier. The final benchmark contains 1,636 verified QA pairs (1,341 grounded and 295 unanswerable) after automated answerability, ambiguity, and consistency checks.

\subsection{Question-Type Distribution}

Table~\ref{tab:qa_type_distribution} reports the distribution of
question types across domains.

\begin{table*}[t]
\centering
\small
\setlength{\tabcolsep}{4pt}
\begin{tabular}{lrrr}
\toprule
\textbf{Domain} &
\textbf{Grounded} &
\textbf{Unanswerable} &
\textbf{Total} \\
\midrule
Clinical Trials
& 331 & 60 & 391 \\

Earnings Calls
& 254 & 60 & 314 \\

FDA 510(k)
& 233 & 59 & 292 \\

Commercial Leases
& 302 & 56 & 358 \\

Municipal Minutes
& 221 & 60 & 281 \\
\midrule
\textbf{Total}
& \textbf{1,341} & \textbf{295}
& \textbf{1,636} \\
\bottomrule
\end{tabular}
\caption{Distribution of QA question types across DSB-DG domains}
\label{tab:qa_type_distribution}
\end{table*}

\subsection{Representative QA Examples}

Table~\ref{tab:qa_examples} provides representative benchmark examples
covering different domains and reasoning types.

\begin{table*}[t]
\centering
\scriptsize
\setlength{\tabcolsep}{3pt}
\begin{tabular}{p{1.7cm}p{1.5cm}p{4.3cm}p{4.3cm}}
\toprule
\textbf{Domain} &
\textbf{Type} &
\textbf{Question} &
\textbf{Reference Answer} \\
\midrule
Clinical Trials
& grounded / direct\_lookup
& How many participants were enrolled in this study?
& 20 participants \\

Earnings Calls
& grounded / direct\_lookup
& What were Southern Company's operating revenues for the fourth quarter of 2025?
& \$7.0 billion \\

FDA 510(k)
& grounded / direct\_lookup
& What is the regulation number associated with the \textit{Francisella tularensis} Real-time PCR Assay?
& 21 CFR 866.4000 \\

Commercial Leases
& grounded / multi\_span
& Who are the landlord and tenant in this lease agreement?
& Landlord: Greenfield Realty Partners, LP; Tenant: David Martin \\

Municipal Minutes
& grounded / direct\_lookup
& Who made the motion to approve the February 2025 meeting minutes?
& Alderman James Rodriguez \\
\bottomrule
\end{tabular}
\caption{Representative DSB-DG question--answer pairs.}
\label{tab:qa_examples}
\end{table*}

\subsection{Examples Rejected During Verification}

Table~\ref{tab:qa_filter_examples} illustrates the adversarial filtering
process.

\begin{table*}[t]
\centering
\scriptsize
\setlength{\tabcolsep}{3pt}
\begin{tabular}{p{1.6cm}p{4.2cm}p{2.2cm}p{4.0cm}}
\toprule
\textbf{Domain} &
\textbf{Candidate Question} &
\textbf{Decision} &
\textbf{Reason} \\
\midrule
FDA 510(k)
& What is the intended use of FMD Peripheral Guidewires?
& Rejected
& Insufficient document evidence \\

FDA 510(k)
& Who is the regulatory contact for submissions related to the FMD Peripheral Guide Wire devices?
& Rejected
& Multiple plausible answers \\

FDA 510(k)
& What is the exact submission date of the 510(k) premarket notification?
& Rejected
& ``Unanswerable'' but actually answerable \\

Earnings
& What were Southern Company's operating revenues for the fourth quarter of 2025?
& Retained
& Unique answer explicitly supported by context \\
\bottomrule
\end{tabular}
\caption{Representative retained and rejected QA candidates from the
verification pipeline.}
\label{tab:qa_filter_examples}
\end{table*}


\section{Automatic Grounding Evaluation}
\label{app:judge_validation}

\subsection{Evaluation Schema}

Model responses are evaluated from the system-produced audio that is converted into text transcripts using Whisper-large-v3. The automatic judge receives the document context, question, reference answer, and response transcript.

\begin{table*}[t]
\centering
\small
\begin{tabular}{lp{10.5cm}}
\toprule
\textbf{Field} & \textbf{Operational Definition} \\
\midrule
Factual Accuracy &
Binary judgment indicating whether the response is factually supported by
the supplied document. \\

Hallucination Type &
\textit{None}; \textit{Intrinsic}, when the response contradicts information
explicitly stated in the document; or \textit{Extrinsic}, when the response
introduces unsupported information not contained in the document. \\

No Response &
The model fails to produce a scoreable answer, including silence or
non-answer conversational behavior. \\

False Refusal &
The model refuses to answer despite the required information being
explicitly available in the supplied context. \\

Partial Credit &
Integer \texttt{partial\_credit} $\in\{0,1,2\}$: $2$ = fully
correct; $1$ = partially correct (captures the main point but misses a detail
or hedges unnecessarily); $0$ = wrong, or refused when the question is
answerable. For \texttt{multi\_span} items a floor of $1$ is enforced when the
response addresses the question and the judge's explanation indicates at least
one span was correct. \\

Question Addressed &
Whether the response substantively attempts to answer the user's question. \\
\bottomrule
\end{tabular}
\caption{Structured output schema used by the automatic grounding judge.}
\label{tab:judge_schema}
\end{table*}

\subsection{LLM Judge Configuration}

We use GPT-OSS-120B as the automatic grounding evaluator. Judge inference uses temperature 0.0, maximum output length 512 tokens, and a fixed rate limit of one request per second. Empty transcripts are flagged \texttt{no\_response} without an LLM call, and unparseable outputs fall back to a default all-zero judgment applied to malformed outputs.

\subsection{Judge Prompt}
\label{app:judge_prompt}
The evaluation judge prompt used in our experiments is reproduced below.

\begin{quote}
\small
\ttfamily
[SYSTEM] You are an expert evaluation judge for a
spoken-language document grounding benchmark. Your task is to assess whether a
spoken model response accurately answers a question based solely on the
provided document.\\[2pt]
SCORING RULES:\\
1. factual\_accuracy (0|1): 1 if the core factual claim in the response
matches the ground truth. 0 otherwise.\\
2. partial\_credit (0|1|2): 2=fully correct, 1=partially correct (gets main
point but misses detail or hedges unnecessarily), 0=wrong or refused when
answerable.\\
3. hallucination\_type: ``none'', ``intrinsic'' (contradicts the document),
``extrinsic'' (introduces facts not in document), or ``both''.\\
4. hallucinated\_claims: list any claims not supported by the document.\\
5. appropriate\_refusal (0|1): 1 if the question is unanswerable and the model
correctly refused/expressed uncertainty.\\
6. false\_confidence (0|1): 1 if the model stated something with certainty that
was wrong or not in the document.\\
7. model\_expressed\_uncertainty (bool); 8. response\_addresses\_question
(bool); 9. explanation (1--2 sentences).\\
Be strict but fair.

\vspace{0.5em}
[USER]
Document: \{doc\_text\}

Question: \{question\}
Question Type: \{question\_type\};
Question Subtype: \{question\_subtype\}

Reference Answer: \{ground\_truth\}
Supporting Sentences: \{supporting\_sentences\}

Model Response: \{model\_transcript\}

\vspace{0.5em}
Output a single JSON object with exactly these keys:
factual\_accuracy (0|1), partial\_credit (0|1|2), hallucination\_type
(none|intrinsic|extrinsic|both), hallucinated\_claims ([]),
appropriate\_refusal (0|1), false\_confidence (0|1),
model\_expressed\_uncertainty (bool), response\_addresses\_question (bool),
explanation (string).
\end{quote}

\section{Model Configuration and Reproducibility}
\label{app:model_config}

\subsection{System Overview}

Table~\ref{tab:model_configuration} summarizes the exact configurations
used in our experiments.

\begin{table*}[t]
\centering
\scriptsize
\setlength{\tabcolsep}{3pt}
\begin{tabular}{llllll}
\toprule
\textbf{System} &
\textbf{Checkpoint} &
\textbf{Backbone} &
\textbf{Context Interface} &
\textbf{Precision} &
\textbf{Hardware/API} \\
\midrule

Cascaded
& whisper-1 / gpt-4o-2024-11-20 / tts-1
& GPT-4o
& System prompt
& --
& hosted OpenAI API \\

GPT-Realtime
& gpt-realtime-2025-08-28
& Not Public
& System item
& --
& hosted OpenAI Realtime API \\

MiniCPM-o
& openbmb/MiniCPM-o-4\_5
& MiniCPM-o 4.5
& System prompt
& bf16
& 1$\times$A100-80GB ($\sim$35\,GB) \\

SALMONN
& tsinghua-ee/SALMONN-7B
& Vicuna-7B-v1.5
& System prompt
& fp16
& 1$\times$A100-80GB ($\sim$27\,GB) \\

UV-32B
& fixie-ai/ultravox-v0\_6-qwen-3-32b
& Qwen3-32B
& System prompt
& bf16
& 1$\times$A100-80GB ($\sim$62\,GB) \\

Moshi
& kyutai/moshiko-pytorch-bf16
& Moshi/Helium-7B LM + Mimi
& KV-cache prefill
& bf16
& 1$\times$A100-80GB ($\sim$46\,GB) \\
\bottomrule
\end{tabular}
\caption{Model, inference, and execution configurations used in DSB-DG.}
\label{tab:model_configuration}
\end{table*}

\subsection{Decoding Hyperparameters}

Table \ref{tab:decoding_params} summarizes the decoding hyperparams used in our experiments. 

\begin{table*}[t]
\centering
\small
\setlength{\tabcolsep}{4pt}
\begin{tabular}{lrrrrl}
\toprule
\textbf{System} &
\textbf{Temperature} &
\textbf{Top-$p$} &
\textbf{Max Output} &
\textbf{Seed} \\
\midrule
Cascaded
& 0.1 & default & 300 & -- \\
GPT-Realtime
& 0.6 & N/A & provider default & -- \\
MiniCPM-o
& greedy & -- & 256 & -- \\
SALMONN
& 1.0 & 0.9 & 150 (max\_length) & -- \\
UV-32B
& greedy & -- & 2048 & -- \\
Moshi
& not exposed & not exposed & N/A & -- \\
\bottomrule
\end{tabular}
\caption{Decoding hyperparameters. For parameters not exposed by an API or
architecture, we report ``N/A'' or ``provider default''.}
\label{tab:decoding_params}
\end{table*}

\subsection{Context Injection Across Architectures}
\label{sec:context_injection_appendix}

As diverse speech LM systems expose different conditioning interfaces, we inject identical document content through each model's native or closest available context mechanism. Cascaded, MiniCPM-o, and UV-32B receive the document through system-level text conditioning. GPT-Realtime and Gemini-Live receive the document as a system conversation item before interaction. SALMONN is similarly conditioned through its system prompt because long-context inputs can cause it to transcribe rather than answer incoming speech. Moshi provides no native text-conditioning interface, so we inject document context by pre-filling the language model's KV cache before interaction. These procedures preserve each architecture while providing access to the same source information. Exact prompt templates and model-specific injection procedures are provided in Appendix~\ref{app:model_config}.

\subsection{Context Injection by Architecture}

\paragraph{Cascaded.} Pipeline is 3-stage: OpenAI Whisper, GPT-4o, and OpenAI TTS. This pipeline is only used as a system to be evaluated, not for any generation of dataset.

\paragraph{GPT-Realtime.} We evaluate \texttt{gpt-realtime-2025-08-28} . The document is inserted as a \texttt{conversation.item.create} item with \texttt{role="system"} sent after \texttt{session.updated} and before the user question before interaction. Session configuration uses \texttt{modalities=["audio","text"]}, temperature 0.6, PCM16 @24\,kHz audio.

\paragraph{Gemini-Live.} We evaluate Gemini-Live API using the model's native
audio-to-audio interface. The document is supplied through the session
\texttt{setup.systemInstruction} using the shared preamble+document template, so
grounding context is fixed for the entire session before any interaction. The
setup requests \texttt{responseModalities=["AUDIO"]} with temperature 0.6, and
disables server-side voice-activity detection so that turn boundaries are controlled explicitly. Each question is streamed as PCM16@16kHz audio framed by \texttt{activityStart}/\texttt{activityEnd} signals; the model responds with PCM16 @24kHz audio, which we collect until \texttt{turnComplete}.

\paragraph{MiniCPM-o.} We evaluate \texttt{openbmb/MiniCPM-o-4\_5}. Document context is supplied through the system prompt. Inference uses greedy decoding (\texttt{sampling=False}, max\_new\_tokens 256) on a single A100-80GB.

\paragraph{SALMONN.} We evaluate \texttt{tsinghua-ee/SALMONN-7B}. Document context is supplied through the system prompt using the shared preamble+document template.

\paragraph{UV-32B.} We evaluate \texttt{fixie-ai/ultravox-v0\_6-qwen-3-32b}. Document
context is supplied using a \texttt{\{"role":"system","content":document\}} message together with the audio user turn. Inference uses bf16 on a single A100-80GB, max\_new\_tokens 2048, with $\langle$think$\rangle\ldots\langle$/think$\rangle$ reasoning blocks stripped before scoring.

\paragraph{Moshi.} We evaluate \texttt{kyutai/moshiko-pytorch-bf16}. Since Moshi exposes no native text-conditioning interface, document context is introduced by pre-filling the language model KV cache: for each document token the server encodes a silent audio frame, runs \texttt{lm\_gen.step()}, then overrides the sampled inner-monologue token with the target document token.

\subsection{Context-Injection Prompts}

The exact system/context templates used for each architecture is provided
below.

\begin{quote}
\small\ttfamily
You are a knowledgeable, concise assistant participating in a
spoken conversation. Answer questions clearly and accurately. Keep responses
under 60 words unless detail is explicitly requested. Do not use bullet points
or headers --- respond in natural spoken prose.\\[3pt]
You have been provided the following reference document. Answer all questions
ONLY using information contained in this document. If the answer cannot be
found in the document, say so clearly.\\[3pt]
DOCUMENT:\\
"""\{document\_text\}"""\\[3pt]
\end{quote}

\noindent In all Proactive-Grounding conditions ($C_0$/$C_1$/$C_2$) the document is present from the start; the conditions differ only in whether/how often it is re-injected mid-conversation.

\subsection{Audio Generation and Standardization}

All benchmark user utterances are synthesized using Kokora-TTS (Apache 2.0 license)\footnote{\url{https://github.com/hexgrad/kokoro}}, with \texttt{en-US-JennyNeural} voice at eval time at 16\,kHz mono. Audio is encoded as 16-bit PCM WAV with 16\,kHz sample rate, single channel. We apply 0.3\,s leading and 0.8\,s trailing silence (model adapters may append additional end-of-turn silence, e.g.\ Moshi adds a 1.5\,s tail). Identical audio files are supplied to all systems where supported.

\subsection{Experimental Runs and Reproducibility}

Each benchmark instance is evaluated 3 times per system and we report the median performance. Failed API/inference calls are handled using exponential-backoff retries with checkpointed resumption.Open-weight models are evaluated on a server with 2$\times$NVIDIA A100-SXM4-80GB, CUDA 12.x, Amazon Linux 2023, Python 3.13, per-model virtual environments (torch/transformers versions per Table~\ref{tab:model_configuration}). API experiments are conducted through the hosted OpenAI/Gemini APIs.


\section{Detailed Evaluation Protocols}
\label{app:protocol_details}

\subsection{Context Saturation}

Each evaluation instance is an independent single-turn interaction.
For a QA pair eligible at tier $T_i$, the identical spoken question is
evaluated at every applicable context tier while conversation state is
reset between runs.

\paragraph{Eligibility.} A question first becomes eligible at the single tier it was generated from --- each QA pair is generated against one specific tiered document and is evaluated only at that tier. Across all five tiers, no questions are shared across every tier, while each question is evaluated at every tier from its minimum eligible tier through T5. The number of QA whose \emph{minimum} eligible tier is $T_1$--$T_5$ is 663/190/317/298/168 respectively; the number of \emph{evaluated turns} per tier (all eligible questions at that tier) is larger and grows with tier. Tiers are content-preserving nested samples.

\paragraph{Session Reset.} Between single-turn evaluations, model/session state is reset using a fully stateless per-turn protocol: each turn re-sends the system/document context; GPT-Realtime opens a fresh WebSocket session per turn; the Moshi and local HTTP servers are stateless across turns.

\paragraph{Scoring and Aggregation.} GA and HR are computed as micro-averages over scored turns (GA = fraction with \texttt{factual\_accuracy}=1; HR = fraction with a non-empty transcript scored \texttt{factual\_accuracy}=0). Domain-level statistics use the same per-turn micro-average restricted to each domain.

\subsection{Grounding Decay}
\label{app:grounding_decay_protocol}

\paragraph{Conversation Construction.}
We construct 20 controlled conversations, with four conversations for each domain. Each conversation contains 17 turns and follows a probe--distractor--repeat structure (see Table \ref{tab:decay_schedule}).

\begin{table}[t]
\centering
\small
\begin{tabular}{ll}
\toprule
\textbf{Turn(s)} & \textbf{Content} \\
\midrule
2--4 & Early document probes ($A$) \\
5--6 & Domain-relevant distractors \\
7--9 & Intermediate probes ($B$) \\
10--12 & Domain-relevant distractors \\
13--15 & Repeated probes ($A'$) \\
\bottomrule
\end{tabular}
\caption{Turn structure of the Grounding Decay protocol (0-indexed turn numbers). Turns 0--1 are un-probed opening turns and turn 16 is a closing turn; the probe/distractor layout is identical across all 20 scripts. Each conversation is 17 turns.}
\label{tab:decay_schedule}
\end{table}

\paragraph{Probe Construction.}
Each conversation contains 3 $A$ probes,
3 $B$ probes, and 3 repeated $A'$ probes.
$A'$ probes are verbatim-identical to the $A$ probes --- the same
question strings, re-asked after intervening dialogue, so changes in accuracy reflect conversational
progression rather than question variation.
Intermediate $B$ probes target different facts in the same
document, so mid-dialogue grounding can be measured independently of the
$A$/$A'$ facts.

\paragraph{Distractors.}
Distractor turns are generated using the conversation-script generator and appear as ``discussion'' turns. They are
constrained to (i) remain semantically related to the professional domain,
(ii) avoid requiring the facts queried by $A/A'$, and (iii) avoid revealing
the answers to later probes. Distractors are validated using
generation-time constraints rather than a separate post-hoc validator.

\paragraph{GDS Aggregation.}
For conversation $c$,
\[
\mathrm{GDS}_c =
\mathrm{GA}_c(A')-\mathrm{GA}_c(A).
\]
System-level GDS is computed using
the unweighted mean of $\mathrm{GDS}_c$ across the 20
conversations (all six systems are run over the same 20 scripts); GA within a
probe group is the fraction of that group's probe turns scored
\texttt{factual\_accuracy}=1.

\subsection{Representative Grounding-Decay Conversation}

Table~\ref{tab:decay_example} provides an abbreviated representative
conversation. The released benchmark contains the complete scripts.

\begin{table*}[t]
\centering
\scriptsize
\setlength{\tabcolsep}{3pt}
\begin{tabular}{clp{8.2cm}}
\toprule
\textbf{Turn} & \textbf{Type} & \textbf{User Utterance} \\
\midrule
2 & Probe $A$ & What is the primary purpose of the study? \\
4 & Probe $A$ & What imaging technique is considered the gold standard in this study? \\
6 & Distractor & Do you think telemedicine is going to completely replace in-person visits someday? \\
7 & Probe $B$ & What is the significance of an AUC value of 0.5? \\
11 & Distractor & What are your thoughts on alternative medicine? Do you think it can really work alongside clinical treatments? \\
13 & Probe $A'$ & What are the key takeaways from the study? \\
\bottomrule
\end{tabular}
\caption{Representative probe--distractor--repeat conversation used in
Experiment B (abbreviated from \texttt{clinical\_001}; 0-indexed turn
numbers). The turn-13 $A'$ probe repeats the turn-2 $A$ probe semantically.}
\label{tab:decay_example}
\end{table*}

\subsection{Proactive Grounding}
\label{app:proactive_grounding_protocol}

Each Proactive Grounding conversation reuses the corresponding 17-turn
Experiment~B script, with nine document-grounded probes (three $A$, three $B$,
and three repeated $A'$). Identical conversation scripts and factual probes
are used across $C_0$, $C_1$, and $C_2$; only the refresh schedule changes.

\paragraph{Refresh Schedule.}

\begin{table*}[t]
\centering
\small
\begin{tabular}{lll}
\toprule
\textbf{Condition} &
\textbf{Initial Context} &
\textbf{Refresh Turn(s)} \\
\midrule
$C_0$ & T3 document (at start) & None \\
$C_1$ & T3 document (at start) & one, before the late ($A'$) probe group \\
$C_2$ & T3 document (at start) & two, before the mid ($B$) and late ($A'$) probe groups \\
\bottomrule
\end{tabular}
\caption{Context schedule for the Proactive Grounding protocol. All three
conditions provide the T3 document at the start; they differ in the number of
mid-conversation document re-injections (refreshes): $C_0$ none, $C_1$ one,
$C_2$ two.}
\label{tab:refresh_schedule}
\end{table*}

The factual probes are the same probe turns used in Experiment B (the early
$A$, mid $B$, and repeated late $A'$ groups), held fixed across $C_0$/$C_1$/$C_2$;
only the refresh schedule changes. A refresh re-injects the T3 document using
each model's native context mechanism (Appendix~\ref{app:model_config})
immediately before the indicated probe group. No other conversation state is
altered between conditions.

\section{Additional Results}
\label{app:additional_results}

\begin{table}[t]
\centering
\small
\setlength{\tabcolsep}{4pt}
\begin{tabular}{lccccc}
\toprule
\textbf{Model} &
\textbf{T1} &
\textbf{T2} &
\textbf{T3} &
\textbf{T4} &
\textbf{T5} \\
\midrule
Cascaded     & 12.2 & 10.6 & 9.5 & 9.5 & 12.3 \\
GPT-Realtime & 11.4 & 14.2 & 11.8 & 12.1 & 14.8 \\
Gemini-Live  & 13.3 & 13.4 & 12.7 & 10.5 & 13.7 \\
MiniCPM-o    & 13.9 & 12.7 & 12.3 & 13.1 & 25.0 \\
UV-32B       & 13.0 & 11.6 & 11.4 & 10.5 & 12.7 \\
SALMONN      & 22.6 & 24.4 & 30.1 & 29.8 & 36.5 \\
Moshi        & 83.4 & 77.1 & 79.2 & 82.7 & 82.0 \\
\bottomrule
\end{tabular}
\caption{Hallucination Rate (\%) across context tiers (intrinsic $+$ extrinsic,
over scoreable turns). MiniCPM-o's HR nearly doubles at T5 (12.3$\to$25.0), mirroring its T5 accuracy drop; SALMONN's HR rises steadily with context. Gemini-Live's HR stays flat and low across tiers (10.5--13.7\%), with no saturation-driven rise. Moshi's high HR reflects fluent but ungrounded speech.}
\label{tab:appendix_hr}
\end{table}

 \paragraph{Hallucinations.} Table~\ref{tab:halluc_decomp} decomposes the Hallucination Rate into \emph{intrinsic} (contradicting the document) and \emph{extrinsic} (unsupported additions) errors, and adds the false-confidence rate and the normalized partial-credit score (0--2). Across \emph{every} model, grounding errors are dominated by \textbf{intrinsic} hallucinations (contradicting the supplied document) rather than \textbf{extrinsic} ones (inventing outside facts). Models are not ``making things up'' so much as \emph{misreading the document in front of them}, and they do so \emph{confidently} --- the false-confidence rate tracks HR closely. Gemini-Live is no exception (10.0\% intrinsic vs.\ 2.6\% extrinsic). This argues that grounding interventions should target document \emph{comprehension/attention}, not external-knowledge suppression.

\begin{table}[t]
\centering
\small
\setlength{\tabcolsep}{4pt}
\begin{tabular}{lrrrrr}
\toprule
\textbf{Model} & \textbf{HR} &
\textbf{Intr.} & \textbf{Extr.} \\
\midrule
Cascaded     & 10.8 & 8.2  & 2.5 \\
MiniCPM-o    & 16.7 & 12.6 & 4.0 \\
GPT-Realtime & 13.1 & 11.4 & 1.7 \\
Gemini-Live  & 12.6 & 10.0 & 2.6 \\
UV-32B       & 11.6 & 10.0 & 1.6 \\
SALMONN      & 29.5 & 27.7 & 1.8 \\
Moshi        & 81.1 & 63.3 & 17.8 \\
\bottomrule
\end{tabular}
\caption{Hallucination decomposition (\%), pooled over tiers T1--T5 (grounded/
answered turns). \emph{Intr.}/\emph{Extr.} are intrinsic/extrinsic hallucination
rates; Errors are overwhelmingly \emph{intrinsic} (misreading the document) rather than \emph{extrinsic} (inventing facts); Moshi is degenerate (GA$=$3.6\%, fluent but ungrounded).}
\label{tab:halluc_decomp}
\end{table}

\begin{table}[t]
\centering
\small
\setlength{\tabcolsep}{5pt}
\begin{tabular}{lrrr}
\toprule
\textbf{Model} & \textbf{NRR (\%) $\uparrow$} & \textbf{FCR (\%) $\downarrow$} \\
\midrule
GPT-Realtime & 86.6  & 12.3 \\
Gemini-Live  & 85.6  & 13.2 \\
Cascaded     & 84.6  & 14.2 \\
UV-32B       & 80.3  & 19.0 \\
MiniCPM-o    & 77.2  & 21.2 \\
SALMONN      & 61.7  & 31.4 \\
Moshi        & 12.5  & 85.7 \\
\bottomrule
\end{tabular}
\caption{Behaviour on \emph{unanswerable} questions: NRR = appropriate-refusal rate (higher is better), FCR = false-confidence rate (lower is better). NRR/FCR are computed from the judge fields \texttt{appropriate\_refusal} and \texttt{false\_confidence} over all unanswerable turns, pooled across tiers T1--T5.}
\label{tab:abstention}
\end{table}

\paragraph{Abstention on Unanswerable Questions.} The strong systems refuse unanswerable questions well (GPT-Realtime 86.6\%, Gemini-Live 85.6\%, Cascaded 84.6\%, UV-32B 80.3\%, MiniCPM-o 77.2\%), whereas SALMONN answers \emph{anyway} roughly 38\% of the time, often with false confidence (FCR 31.4\%), and Moshi almost never abstains (NRR 12.5\%, FCR 85.7\%). Appropriate refusal broadly tracks grounding quality here, yet remains a distinct axis: GPT-Realtime abstains slightly better than the Cascaded oracle despite lower single-turn QA accuracy.

\begin{table}[t]
\centering
\small
\setlength{\tabcolsep}{4pt}
\begin{tabular}{lrrr@{\hskip 10pt}rrr}
\toprule
& \multicolumn{3}{c}{\textbf{Difficulty}} &
  \multicolumn{3}{c}{\textbf{Evidence position}} \\
\cmidrule(lr){2-4}\cmidrule(l){5-7}
\textbf{Model} & Easy & Med. & Hard & Early & Mid & Late \\
\midrule
Cascaded     & 91.1 & 87.4 & 81.9 & 89.5 & 85.5 & 91.2 \\
GPT-Realtime & 89.7 & 86.7 & 68.1 & 84.2 & 87.0 & 88.6 \\
Gemini-Live  & 87.6 & 89.1 & 82.6 & 83.4 & 88.0 & 90.4 \\
MiniCPM-o    & 86.3 & 80.4 & 73.5 & 85.0 & 80.7 & 84.3 \\
UV-32B       & 81.5 & 76.4 & 72.7 & 79.1 & 73.0 & 83.2 \\
SALMONN      & 66.4 & 51.2 & 44.0 & 62.3 & 60.7 & 59.2 \\
\bottomrule
\end{tabular}
\caption{Grounding accuracy (\%) by annotated question difficulty and by the
position of the supporting evidence in the document.
\texttt{difficulty} is a QA-metadata field; evidence position is bucketed
by the character offset of the first supporting sentence in the one-third splits.}
\label{tab:difficulty_position}
\end{table}

\paragraph{Sensitivity to Question Difficulty and Evidence Position.} Nearly every model degrades monotonically from easy to hard questions, and the gap widens for the weaker models (GPT-Realtime $89.7\!\to\!68.1$; SALMONN $66.4\!\to\!44.0$); Gemini-Live is the exception, peaking on medium questions (easy 87.6, medium 89.1, hard 82.6) before the expected drop on hard. Evidence position has a smaller, less consistent effect: GPT-Realtime dips on mid/late evidence but most other models, like UV-32B, are essentially position-invariant, indicating that question difficulty, not evidence placement, is the dominant factor at short context lengths; Gemini-Live is a mild outlier in the opposite direction, improving as the evidence moves later (early 83.4 $\to$ late 90.4).

\begin{table}[t]
\centering
\small
\setlength{\tabcolsep}{6pt}
\begin{tabular}{lrrr}
\toprule
\textbf{Model} & \textbf{Median} & \textbf{Mean} & \textbf{p95} \\
\midrule
Cascaded     & 14 & 17.1 & 43 \\
GPT-Realtime & 18 & 18.7 & 33 \\
Gemini-Live  & 13 & 14.4 & 30 \\
MiniCPM-o    & 14 & 28.4 & 75 \\
UV-32B       & 14 & 17.5 & 43 \\
SALMONN      & 15 & 24.5 & 94 \\
Moshi        & 29 & 30.3 & 54 \\
\bottomrule
\end{tabular}
\caption{Response length in whitespace tokens (transcripts;
$\langle$think$\rangle$ blocks stripped and \texttt{/no\_think} used for UV-32B).
With reasoning suppressed, UV-32B is now as concise as the other
systems (median 14); Gemini-Live is the most concise (median 13); Moshi is the most verbose (median 29) despite not
grounding, and MiniCPM-o/SALMONN have the longest tails.}
\label{tab:verbosity}
\end{table}

\section{Qualitative Analysis}
\label{app:qualitative}

We complement aggregate metrics with representative transcripts from the model runs. The examples below are selected to illustrate grounded retrieval, hallucination, context-capacity failure, multi-turn retention, and the effect of context refresh. Model responses are the normalized transcripts used for evaluation.

\begin{examplebox}{Grounded retrieval at long context --- Cascaded}
\textbf{Document evidence.} \emph{If Tenant fails to pay rent within 3 days after the due date, Tenant shall pay a late fee of \$100.}

\textbf{Question.} What is the late fee for rent payments made after three days past the due date?

\textbf{Reference.} \$100.

\textbf{Response.} \emph{``The late fee for rent payments made more than three days after the due date is \$100.''}

\textbf{Takeaway.} The cascaded system retrieves the exact numerical fact at the longest context tier, illustrating its robustness under contextual load.
\end{examplebox}

\begin{examplebox}{Extrinsic hallucination on a correct core answer --- UV-32B}
\textbf{Document evidence.} \emph{Mayor Laura Peterson: ``Any opposed? Any abstentions? Hearing none, the motion carries. The minutes from February 11, 2025, are approved.''}

\textbf{Question.} What was the outcome of the vote on the February 2025 meeting minutes?

\textbf{Reference.} The motion carries.

\textbf{Response.} \emph{``The minutes from the February 11, 2025, meeting were approved unanimously.''}

\textbf{Takeaway.} The core outcome is correct, but the response adds ``unanimously,'' which is not stated in the document. This illustrates an extrinsic hallucination layered onto an otherwise grounded answer.
\end{examplebox}

\begin{examplebox}{False confidence on an unanswerable question --- GPT-Realtime}
\textbf{Question.} What is the total number of arms in this clinical trial?

\textbf{Reference.} Unanswerable from the supplied document.

\textbf{Response.} \emph{``There are 15 study arms described in the document.''}

\textbf{Takeaway.} Even a strong system can occasionally fabricate a specific answer instead of abstaining when the requested fact is absent from the document.
\end{examplebox}

\begin{examplebox}{Context-capacity cliff --- SALMONN}
\textbf{Document evidence.} \emph{Southern Company also reported full-year 2025 earnings of \$4.3 billion, or \$3.94 per share, compared with \$4.4 billion, or \$4.02 per share, in 2024.}

\textbf{Question.} Did Southern Company have a higher or lower net income in 2025 compared to 2024?

\textbf{Reference.} Lower.

\textbf{T3 response ($\approx$2k tokens).} \emph{``Southern Company had a lower net income in 2025 compared to 2024.''}

\textbf{T4 response ($\approx$4k tokens).} \emph{``A point 2 a point 2 a point 2. Dot a. Point 2 a. Point 2 a. Point 2. 2. 2. 2. 2. 22. 22. 2. 2. Minus 2. 8. 8. 8.''}

\textbf{Takeaway.} SALMONN answers correctly at T3 but degenerates at T4, matching the abrupt GA drop from 70.3\% to 37.2\% in Context Saturation Protocol.
\end{examplebox}

\begin{examplebox}{Measurable multi-turn decay --- UV-32B}
\textbf{Document evidence.} \emph{An unreversed 2-point decline is any decline of 2 points or more that had not reversed to a 1-point decline (or better) at last recorded observation.}

\textbf{Question.} What does the term ``unreversed 2-point decline'' mean in the context of the study?

\textbf{Reference.} A decline of 2 points or more that had not reversed to a 1-point decline or better at the last recorded observation.

\textbf{Early probe $A$.} \emph{``An `unreversed 2-point decline' refers to a decrease of 2 points or more in the motor-language (ML) score that has not improved back to a 1-point decline or better by the final recorded observation.''}

\textbf{Repeated probe $A'$.} \emph{``An `unreversed 2-point decline' refers to a reduction of 2 points or more in the motor-language (ML) score that has not recovered to within 1 point of the original level or better\ldots''}

\textbf{Takeaway.} The repeated answer subtly changes the document definition and is judged unsupported.
\end{examplebox}

\begin{examplebox}{Stable retention --- Cascaded}
\textbf{Document evidence.} \emph{Primary Purpose: DIAGNOSTIC.}

\textbf{Question.} What is the primary purpose of the study?

\textbf{Reference.} DIAGNOSTIC.

\textbf{Early probe $A$.} \emph{``The primary purpose of the study is diagnostic.''}

\textbf{Repeated probe $A'$.} \emph{``The primary purpose of the study is diagnostic.''}

\textbf{Takeaway.} The cascaded system preserves the same grounded fact after intervening dialogue, consistent with its aggregate GDS of 0.0 pp.
\end{examplebox}

\begin{examplebox}{Turn-to-turn instability rather than monotonic decay --- SALMONN}
\textbf{Document evidence.} \emph{NCT Identifier: NCT04424290.}

\textbf{Question.} What is the NCT Identifier for the HORNBILL study?

\textbf{Reference.} NCT04424290.

\textbf{Early probe $A$.} \emph{``The purpose of the study described in the document is to test different doses of BI 764524 in patients who have had laser treatment for a type of diabetic eye disease\ldots''}

\textbf{Repeated probe $A'$.} \emph{``The NCT Identifier for the HORNBILL study is NCT04424290.''}

\textbf{Takeaway.} SALMONN fails the early probe but answers the repeat correctly. Its positive GDS therefore reflects turn-level instability around weaker absolute grounding rather than improving memory.
\end{examplebox}

\section{Detailed Latency Analysis}
\label{app:latency_analysis}

\subsection{FTED Measurement}

First-Token Emission Delay (FTED) is measured from the end of the user audio segment to the first emitted model audio unit. Timing is recorded using wall-clock timers in the streaming evaluation harness. Measurements include network transit time,
include audio encoding/decoding, and exclude queuing or connection-establishment overhead. Persistent API sessions are not used for Context Saturation protocol --- each
turn opens a fresh stateless session. We report p50, mean, and p95 across turns.

\begin{table*}[t]
\centering
\scriptsize
\setlength{\tabcolsep}{3pt}
\begin{tabular}{llrrr}
\toprule
\textbf{Model} &
\textbf{Tier} &
\textbf{p50 (ms)} &
\textbf{Mean (ms)} &
\textbf{p95 (ms)} \\
\midrule
Moshi        & T1 & 229 & 241 & 768 \\
Moshi        & T2 & 229 & 254 & 781 \\
Moshi        & T3 & 229 & 267 & 895 \\
Moshi        & T4 & 229 & 236 & 612 \\
Moshi        & T5 & 229 & 255 & 712 \\
\midrule
GPT-Realtime & T1 & 1210 & 1338 & 1851 \\
GPT-Realtime & T2 & 1221 & 1315 & 1675 \\
GPT-Realtime & T3 & 1209 & 1300 & 1728 \\
GPT-Realtime & T4 & 1212 & 1310 & 1732 \\
GPT-Realtime & T5 & 1214 & 1308 & 1817 \\
\midrule
SALMONN      & T1 & 1981 & 2352 & 5110 \\
SALMONN      & T2 & 2476 & 2987 & 6657 \\
SALMONN      & T3 & 3783 & 4307 & 8956 \\
SALMONN      & T4 & 7129 & 16205 & 48850 \\
SALMONN      & T5 & 4593 & 8670 & 32058 \\
\midrule
Cascaded     & T1 & 1228 & 9366 & 61440 \\
Cascaded     & T2 & 1414 & 8845 & 61457 \\
Cascaded     & T3 & 1435 & 8263 & 61594 \\
Cascaded     & T4 & 1495 & 7516 & 61561 \\
Cascaded     & T5 & 1588 & 7669 & 61653 \\
\midrule
MiniCPM-o    & T1 & 1412 & 1598 & 3400 \\
MiniCPM-o    & T2 & 1602 & 1925 & 4180 \\
MiniCPM-o    & T3 & 2011 & 2579 & 5799 \\
MiniCPM-o    & T4 & 2774 & 3531 & 8691 \\
MiniCPM-o    & T5 & 4068 & 9494 & 44204 \\
\midrule
UV-32B       & T1 & 6832 & 8405 & 19608 \\
UV-32B       & T2 & 7859 & 8620 & 17918 \\
UV-32B       & T3 & 9724 & 10291 & 19229 \\
UV-32B       & T4 & 11804 & 12536 & 23381 \\
UV-32B       & T5 & 12639 & 13291 & 26103 \\
\midrule
Gemini-Live  & T1 & 6395 & 6573 & 8865 \\
Gemini-Live  & T2 & 6583 & 6751 & 9095 \\
Gemini-Live  & T3 & 6761 & 6904 & 9318 \\
Gemini-Live  & T4 & 6931 & 7026 & 9505 \\
Gemini-Live  & T5 & 7034 & 7166 & 9624 \\
\bottomrule
\end{tabular}
\caption{FTED statistics by model and context tier.
Per-tier p50/mean/p95 recomputed from the per-instance
\texttt{fted\_ms} fields (Exp A). Latency grows with context for MiniCPM-o,
SALMONN, and UV-32B, while Moshi and GPT-Realtime are essentially
context-invariant; SALMONN's high T4 p95 is driven by large-context stalls, and
the cascaded mean/p95 are inflated by a long tail from occasional LLM/API
stalls. Gemini-Live has a high but nearly context-invariant latency floor
(p50 6.4--7.0\,s). UV-32B is not real-time capable.}
\label{tab:latency_by_tier}
\end{table*}

\paragraph{Latency Measurement Environment.} Open-weight latency measurements are collected on NVIDIA A100-SXM4-80GB GPUs under CUDA 12.x, using each model's native PyTorch/transformers inference stack (per-model virtual environments). To reduce transient network effects, models/servers are loaded once and kept warm and turns run sequentially; Whisper-large-v3 ASR of the model's spoken response is run on GPU where a card is free (otherwise CPU) and is not counted in FTED.

\begin{tcolorbox}[
  enhanced,
  breakable,
  colback=green!3,
  colframe=green!45!black,
  boxrule=0.5pt,
  arc=2pt,
  left=6pt,right=6pt,top=5pt,bottom=5pt,
  title=\textbf{Cascaded --- grounded numerical lookup at long context},
  fonttitle=\small
]
\textbf{Document context.} ``If Tenant fails to pay rent within 3 days after the due date, Tenant shall pay a late fee of \$100.''

\textbf{Question.} What is the late fee for rent payments made after three days past the due date?

\textbf{Reference answer.} \$100.

\textbf{Model response.} ``The late fee for rent payments made more than three days after the due date is \$100.''

\textbf{Observation.} The response retrieves the exact documented figure at the longest context tier, illustrating robust grounding under large context.
\end{tcolorbox}

\begin{tcolorbox}[
  enhanced,
  breakable,
  colback=green!3,
  colframe=green!45!black,
  boxrule=0.5pt,
  arc=2pt,
  left=6pt,right=6pt,top=5pt,bottom=5pt,
  title=\textbf{GPT-Realtime --- grounded proper-noun retrieval at long context},
  fonttitle=\small
]
\textbf{Document context.} ``c/o Kelliann Payne, Partner, Hogan Lovells US LLP.''

\textbf{Question.} What organization is Kelliann Payne associated with?

\textbf{Reference answer.} Hogan Lovells US LLP.

\textbf{Model response.} ``Kellyanne Payne is associated with Hogan Levels US LLP, where she's listed as a partner.''

\textbf{Observation.} The audio transcript contains minor proper-noun corruption, but the underlying organization is correctly grounded.
\end{tcolorbox}

\begin{tcolorbox}[
  enhanced,
  breakable,
  colback=green!3,
  colframe=green!45!black,
  boxrule=0.5pt,
  arc=2pt,
  left=6pt,right=6pt,top=5pt,bottom=5pt,
  title=\textbf{MiniCPM-o --- grounded entity lookup},
  fonttitle=\small
]
\textbf{Document context.} ``Bynum Maximilian, International Business Director.''

\textbf{Question.} Who is the International Business Director mentioned in the document?

\textbf{Reference answer.} Bynum Maximilian.

\textbf{Model response.} ``The international business director mentioned in the document is Bynum Maximilian.''

\textbf{Observation.} A straightforward entity lookup is answered correctly at short context.
\end{tcolorbox}

\begin{tcolorbox}[
  enhanced,
  breakable,
  colback=green!3,
  colframe=green!45!black,
  boxrule=0.5pt,
  arc=2pt,
  left=6pt,right=6pt,top=5pt,bottom=5pt,
  title=\textbf{UV-32B --- grounded numerical retrieval},
  fonttitle=\small
]
\textbf{Document context.} ``... evaluate the safety, tolerability, and efficacy of BMN 190 intracerebroventricular (ICV) administration every other week (qow) for a period of 144 weeks, in patients with CLN2.''

\textbf{Question.} What is the duration of BMN 190 administration in this study?

\textbf{Reference answer.} 144 weeks.

\textbf{Model response.} ``BMN 190 was administered every other week for a period of 144 weeks in this study.''

\textbf{Observation.} The model provides a concise, correctly grounded numerical answer.
\end{tcolorbox}

\begin{tcolorbox}[
  enhanced,
  breakable,
  colback=green!3,
  colframe=green!45!black,
  boxrule=0.5pt,
  arc=2pt,
  left=6pt,right=6pt,top=5pt,bottom=5pt,
  title=\textbf{SALMONN --- grounded clause interpretation},
  fonttitle=\small
]
\textbf{Document context.} ``A positive rate of decline means that the subject declined, a negative rate of decline means that the subject improved.''

\textbf{Question.} What is the meaning of a negative rate of decline in the ML score?

\textbf{Reference answer.} A negative rate of decline means that the subject improved.

\textbf{Model response.} ``A negative rate of decline means that the subject improved.''

\textbf{Observation.} SALMONN can ground short contextual queries before the capacity failure observed at longer context.
\end{tcolorbox}

\begin{tcolorbox}[
  enhanced,
  breakable,
  colback=red!3,
  colframe=red!45!black,
  boxrule=0.5pt,
  arc=2pt,
  left=6pt,right=6pt,top=5pt,bottom=5pt,
  title=\textbf{Cascaded --- intrinsic hallucination on a drug name},
  fonttitle=\small
]
\textbf{Document context.} ``Study Title: A Trial of 10 and 30 mg Doses of CVL-231 (Emraclidine) in Participants With Schizophrenia.''

\textbf{Question.} What is the name of the drug being studied in this trial?

\textbf{Reference answer.} CVL-231 (Emraclidine).

\textbf{Model response.} ``The drug being studied in this trial is called CVL-231, Imraclidine.''

\textbf{Observation.} The response corrupts the documented drug name, yielding an intrinsic factual error despite the otherwise strong cascaded baseline.
\end{tcolorbox}

\begin{tcolorbox}[
  enhanced,
  breakable,
  colback=red!3,
  colframe=red!45!black,
  boxrule=0.5pt,
  arc=2pt,
  left=6pt,right=6pt,top=5pt,bottom=5pt,
  title=\textbf{MiniCPM-o --- numerical grounding error},
  fonttitle=\small
]
\textbf{Document context.} ``N901 is the number of 901 subjects matched to 203 subjects (i.e. 29) and N203 is the number of 203 subjects who had matches (i.e. 12).''

\textbf{Question.} How many evaluable participants were in the comparator population?

\textbf{Reference answer.} 29.

\textbf{Model response.} ``There were 203 evaluable participants in the comparator population.''

\textbf{Observation.} The model selects a salient nearby number instead of the parenthetical answer required by the document.
\end{tcolorbox}

\begin{tcolorbox}[
  enhanced,
  breakable,
  colback=red!3,
  colframe=red!45!black,
  boxrule=0.5pt,
  arc=2pt,
  left=6pt,right=6pt,top=5pt,bottom=5pt,
  title=\textbf{UV-32B --- extrinsic addition on a correct answer},
  fonttitle=\small
]
\textbf{Document context.} ``Mayor Laura Peterson: `Any opposed? Any abstentions? Hearing none, the motion carries. The minutes from February 11, 2025, are approved.'''"

\textbf{Question.} What was the outcome of the vote on the February 2025 meeting minutes?

\textbf{Reference answer.} The motion carries.

\textbf{Model response.} ``The minutes from the February 11, 2025, meeting were approved unanimously.''

\textbf{Observation.} The core answer is correct, but the response adds ``unanimously,'' which is not stated in the document.
\end{tcolorbox}

\begin{tcolorbox}[
  enhanced,
  breakable,
  colback=red!3,
  colframe=red!45!black,
  boxrule=0.5pt,
  arc=2pt,
  left=6pt,right=6pt,top=5pt,bottom=5pt,
  title=\textbf{GPT-Realtime --- false confidence on an unanswerable question},
  fonttitle=\small
]
\textbf{Question.} What is the total number of arms in this clinical trial?

\textbf{Reference.} Unanswerable: the total number of study arms is not stated in the document.

\textbf{Model response.} ``There are 15 study arms described in the document.''

\textbf{Observation.} The model fabricates a specific count rather than abstaining when the document does not provide the requested information.
\end{tcolorbox}

\begin{tcolorbox}[
  enhanced,
  breakable,
  colback=red!3,
  colframe=red!45!black,
  boxrule=0.5pt,
  arc=2pt,
  left=6pt,right=6pt,top=5pt,bottom=5pt,
  title=\textbf{Cascaded --- false confidence on an unanswerable question},
  fonttitle=\small
]
\textbf{Question.} How many residents spoke in favor of the zoning variance during public comments?

\textbf{Reference.} Unanswerable: the number of residents who spoke in favor is not stated.

\textbf{Model response.} ``Two residents spoke in favor of the zoning variants during public comments.''

\textbf{Observation.} Even the strong cascaded baseline can over-answer an unanswerable question.
\end{tcolorbox}

\begin{tcolorbox}[
  enhanced,
  breakable,
  colback=green!3,
  colframe=green!45!black,
  boxrule=0.5pt,
  arc=2pt,
  left=6pt,right=6pt,top=5pt,bottom=5pt,
  title=\textbf{GPT-Realtime --- calibrated refusal},
  fonttitle=\small
]
\textbf{Question.} What was the total amount of money appropriated by Council Bill 121182?

\textbf{Reference.} Unanswerable: the document mentions that the bill appropriates money to pay certain claims but does not state the total amount.

\textbf{Model response.} ``I don't have that information from the document. It mentions that Council Bill 121182 appropriates money to pay certain claims, but it doesn't state the total amount.''

\textbf{Observation.} The response correctly refuses while identifying why the supplied document is insufficient.
\end{tcolorbox}

\begin{tcolorbox}[
  enhanced,
  breakable,
  colback=red!3,
  colframe=red!45!black,
  boxrule=0.5pt,
  arc=2pt,
  left=6pt,right=6pt,top=5pt,bottom=5pt,
  title=\textbf{Moshi --- fluent but document-ungrounded speech},
  fonttitle=\small
]
\textbf{Document context.} ``One (1) parking space shall be provided to Tenant for an additional monthly fee of \$50, payable together with the monthly rent.''

\textbf{Question.} How many parking spaces are included in the lease, and what is the cost?

\textbf{Reference answer.} One parking space for an additional \$50 monthly fee.

\textbf{Model response.} ``Poly is just fine. It's the dark of the holidays. There are 18 poly spaces in the league. The cost of a poly space varies depending on the team, but it is usually around \$250,000.''

\textbf{Observation.} The response is fluent but unrelated to the supplied document, illustrating the gap between conversational fluency and factual grounding.
\end{tcolorbox}

\begin{tcolorbox}[
  enhanced,
  breakable,
  colback=orange!5,
  colframe=orange!65!black,
  boxrule=0.5pt,
  arc=2pt,
  left=6pt,right=6pt,top=5pt,bottom=5pt,
  title=\textbf{SALMONN --- context-capacity cliff},
  fonttitle=\small
]
\textbf{Question.} Did Southern Company have a higher or lower net income in 2025 compared to 2024?

\textbf{Reference answer.} Lower.

\textbf{Response at T3 ($\approx$2k tokens).} ``Southern Company had a lower net income in 2025 compared to 2024.''

\textbf{Response at T4 ($\approx$4k tokens).} ``A point 2 a point 2 a point 2. Dot a. Point 2 a. Point 2 a. Point 2. 2. 2. 2. 2. 22. 22. 2. 2. Minus 2. 8. 8. 8.''

\textbf{Observation.} The same question is answered correctly at T3 but degenerates into token-repetition output at T4, illustrating the abrupt context-capacity failure.
\end{tcolorbox}

\begin{tcolorbox}[
  enhanced,
  breakable,
  colback=red!3,
  colframe=red!45!black,
  boxrule=0.5pt,
  arc=2pt,
  left=6pt,right=6pt,top=5pt,bottom=5pt,
  title=\textbf{UV-32B --- multi-turn grounding decay},
  fonttitle=\small
]
\textbf{Question.} What is the trial's unique identifier?

\textbf{Reference answer.} NCT04424290.

\textbf{Early probe $A$.} ``The identifier for the HORNBILL study is NCT04424290.''

\textbf{Late probe $A'$.} The model produces an incomplete reasoning sequence that drifts to the trial's enrollment count and never restates the identifier.

\textbf{Observation.} The model initially accesses the document correctly but fails to recover the same fact after intervening dialogue.
\end{tcolorbox}

\begin{tcolorbox}[
  enhanced,
  breakable,
  colback=green!3,
  colframe=green!45!black,
  boxrule=0.5pt,
  arc=2pt,
  left=6pt,right=6pt,top=5pt,bottom=5pt,
  title=\textbf{Cascaded --- stable grounding across dialogue},
  fonttitle=\small
]
\textbf{Question.} What is the primary purpose of the study?

\textbf{Reference answer.} DIAGNOSTIC.

\textbf{Early probe $A$.} ``The primary purpose of the study is diagnostic.''

\textbf{Late probe $A'$.} ``The primary purpose of the study is diagnostic.''

\textbf{Observation.} The cascaded system reproduces the same grounded fact after intervening dialogue.
\end{tcolorbox}

\begin{tcolorbox}[
  enhanced,
  breakable,
  colback=green!3,
  colframe=green!45!black,
  boxrule=0.5pt,
  arc=2pt,
  left=6pt,right=6pt,top=5pt,bottom=5pt,
  title=\textbf{GPT-Realtime --- stable retention with correct elaboration},
  fonttitle=\small
]
\textbf{Question.} What is the primary purpose of the study?

\textbf{Reference answer.} DIAGNOSTIC.

\textbf{Early probe $A$.} ``The primary purpose of the study is diagnostic.''

\textbf{Late probe $A'$.} ``The primary purpose of the study is diagnostic. It's designed to evaluate and compare imaging-based measures for coronary artery disease.''

\textbf{Observation.} The model retains the core fact and adds supporting detail consistent with the document.
\end{tcolorbox}

\begin{tcolorbox}[
  enhanced,
  breakable,
  colback=green!3,
  colframe=green!45!black,
  boxrule=0.5pt,
  arc=2pt,
  left=6pt,right=6pt,top=5pt,bottom=5pt,
  title=\textbf{MiniCPM-o --- stable grounding across dialogue},
  fonttitle=\small
]
\textbf{Question.} What imaging technique is considered the gold standard in this study?

\textbf{Reference answer.} PET myocardial perfusion imaging (PET-MPI).

\textbf{Early probe $A$.} ``The imaging technique considered the gold standard in this study is PET myocardial perfusion imaging (PET-MPI).''

\textbf{Late probe $A'$.} ``The imaging technique considered the gold standard in this study is PET myocardial perfusion imaging (PET-MPI).''

\textbf{Observation.} The model retains the same document-grounded fact across intervening dialogue.
\end{tcolorbox}

\begin{tcolorbox}[
  enhanced,
  breakable,
  colback=red!3,
  colframe=red!45!black,
  boxrule=0.5pt,
  arc=2pt,
  left=6pt,right=6pt,top=5pt,bottom=5pt,
  title=\textbf{SALMONN --- turn-to-turn grounding instability},
  fonttitle=\small
]
\textbf{Question.} What is the NCT Identifier for the HORNBILL study?

\textbf{Reference answer.} NCT04424290.

\textbf{Early probe $A$.} The model answers an unrelated question about the study rather than providing the identifier.

\textbf{Late probe $A'$.} ``The NCT Identifier for the HORNBILL study is NCT04424290.''

\textbf{Observation.} The model is unstable across turns rather than showing a simple monotonic decay pattern.
\end{tcolorbox}

\begin{tcolorbox}[
  enhanced,
  breakable,
  colback=green!3,
  colframe=green!45!black,
  boxrule=0.5pt,
  arc=2pt,
  left=6pt,right=6pt,top=5pt,bottom=5pt,
  title=\textbf{Moshi --- isolated grounded late probe},
  fonttitle=\small
]
\textbf{Question.} Has Pfizer completed any share repurchases so far in 2024?

\textbf{Reference answer.} No.

\textbf{Early probe $A$.} ``Yes, as of May 20, 2024, Pfizer has completed \$2.5 billion of share repurchases, representing 0.6\% of the outstanding shares.''

\textbf{Late probe $A'$.} ``Pfizer has not completed any share repurchases in 2024.''

\textbf{Observation.} This example illustrates that a non-zero GDS can occur even when absolute grounding remains very low.
\end{tcolorbox}


\begin{tcolorbox}[
  enhanced,
  breakable,
  colback=green!3,
  colframe=green!45!black,
  boxrule=0.5pt,
  arc=2pt,
  left=6pt,right=6pt,top=5pt,bottom=5pt,
  title=\textbf{Gemini-Live --- grounded numerical retrieval at long context},
  fonttitle=\small
]
\textbf{Document context.} ``Our Surestone decking sales increased 44 percent and our traditional wood plastic composite decking increased 35 percent, both from the quarter a year ago.''

\textbf{Question.} How did the Surestone decking sales perform in the fourth quarter of 2025 compared to a year ago?

\textbf{Reference answer.} Increased 44 percent.

\textbf{Model response.} ``Surestone decking sales increased 44 percent in the fourth quarter of 2025 compared to a year ago.''

\textbf{Observation.} At the longest context tier (T5) the model isolates the correct figure among adjacent distractor percentages, illustrating robust numerical grounding under large context.
\end{tcolorbox}

\begin{tcolorbox}[
  enhanced,
  breakable,
  colback=green!3,
  colframe=green!45!black,
  boxrule=0.5pt,
  arc=2pt,
  left=6pt,right=6pt,top=5pt,bottom=5pt,
  title=\textbf{Gemini-Live --- calibrated refusal with justification},
  fonttitle=\small
]
\textbf{Question.} What were the locations of the study sites?

\textbf{Reference.} Unanswerable: the document does not list the locations of the study sites.

\textbf{Model response.} ``I don't have that information, as the document does not list the locations of the study sites.''

\textbf{Observation.} The model correctly abstains and states \emph{why} the supplied document is insufficient, consistent with its high appropriate-refusal rate (NRR $=85.6\%$).
\end{tcolorbox}

\begin{tcolorbox}[
  enhanced,
  breakable,
  colback=green!3,
  colframe=green!45!black,
  boxrule=0.5pt,
  arc=2pt,
  left=6pt,right=6pt,top=5pt,bottom=5pt,
  title=\textbf{Gemini-Live --- stable multi-turn retention},
  fonttitle=\small
]
\textbf{Question.} What were the fourth-quarter 2025 operating revenues?

\textbf{Reference answer.} \$7.0 billion.

\textbf{Early probe $A$.} ``Fourth-quarter 2025 operating revenues were \$7.0 billion.''

\textbf{Late probe $A'$.} ``Fourth-quarter 2025 operating revenues were \$7.0 billion.''

\textbf{Observation.} The model reproduces the identical grounded figure after intervening distractor dialogue, consistent with its near-zero grounding-decay score (GDS $=+0.03$).
\end{tcolorbox}

\begin{tcolorbox}[
  enhanced,
  breakable,
  colback=red!3,
  colframe=red!45!black,
  boxrule=0.5pt,
  arc=2pt,
  left=6pt,right=6pt,top=5pt,bottom=5pt,
  title=\textbf{Gemini-Live --- intrinsic hallucination on a named entity},
  fonttitle=\small
]
\textbf{Document context.} ``Contact Person: Anastasya Melnikov \quad a.melnikov@diagnocat.com''

\textbf{Question.} Who prepared the 510(k) summary for Diagnocat?

\textbf{Reference answer.} Anastasya Melnikov.

\textbf{Model response.} ``The 510(k) summary was prepared by DGNCT LLC.''

\textbf{Observation.} The model substitutes the applicant company for the documented contact person, producing an intrinsic error that contradicts the supplied document rather than inventing an outside fact.
\end{tcolorbox}

\begin{tcolorbox}[
  enhanced,
  breakable,
  colback=red!3,
  colframe=red!45!black,
  boxrule=0.5pt,
  arc=2pt,
  left=6pt,right=6pt,top=5pt,bottom=5pt,
  title=\textbf{Gemini-Live --- extrinsic addition on a correct answer},
  fonttitle=\small
]
\textbf{Document context.} ``Landlord shall not be liable to Tenant for any interruption in utility services caused by circumstances beyond Landlord's control, including but not limited to acts of the utility provider, weather events, or acts of God.''

\textbf{Question.} What is the landlord's responsibility if a utility outage occurs?

\textbf{Reference answer.} Landlord shall not be liable to Tenant for any interruption in utility services caused by circumstances beyond Landlord's control.

\textbf{Model response.} ``Landlord shall not be liable for any interruption in utility services caused by circumstances beyond Landlord's control. The tenant must promptly notify Landlord of any outage.''

\textbf{Observation.} The core clause is grounded correctly, but the response appends a tenant-notification obligation that does not appear in the document --- an extrinsic addition layered onto a correct answer.
\end{tcolorbox}

\begin{tcolorbox}[
  enhanced,
  breakable,
  colback=red!3,
  colframe=red!45!black,
  boxrule=0.5pt,
  arc=2pt,
  left=6pt,right=6pt,top=5pt,bottom=5pt,
  title=\textbf{Gemini-Live --- false confidence on an unanswerable question},
  fonttitle=\small
]
\textbf{Question.} How many people spoke during the public comment section in total?

\textbf{Reference.} Unanswerable: the total number of public speakers is not stated in the document.

\textbf{Model response.} ``According to the document, 25 people spoke during the public comment section.''

\textbf{Observation.} Rather than abstaining, the model fabricates a specific count and attributes it to the document --- the dominant failure mode on the $\sim$14\% of unanswerable questions it does not refuse (FCR $=13.2\%$).
\end{tcolorbox}

\end{document}